\documentclass[runningheads]{llncs}

\usepackage{eccv}

\usepackage{eccvabbrv}

\usepackage{graphicx}
\usepackage{booktabs}
\usepackage{listings}
\usepackage{multirow}

\usepackage[accsupp]{axessibility}  

\usepackage[pagebackref,breaklinks,colorlinks,citecolor=eccvblue]{hyperref}

\usepackage{orcidlink}

\lstdefinestyle{dtypestyle}{ basicstyle=\ttfamily\bfseries\color{blue} }
\lstdefinestyle{cmdstyle}{ basicstyle=\ttfamily\bfseries\color{black} }
\newcommand{\paramstyle}[2]{{\textbf{\texttt{\textcolor{#1}{#2}}}}}

\definecolor{darkgray}{gray}{0.3}
\definecolor{darkgreen}{RGB}{50, 150, 50}
\definecolor{darkred}{RGB}{180, 50, 50}
\definecolor{darkblue}{RGB}{80, 106, 156}
\definecolor{col1}{HTML}{506A88}
\definecolor{col2}{HTML}{2E3D47}
\definecolor{col3}{HTML}{98A9B5}
\definecolor{col4}{HTML}{D19C83}
\definecolor{fom}{RGB}{0,153,139}

\lstdefinelanguage{StructuredLanguage}
{
  basicstyle=\ttfamily\bfseries,
  morekeywords={id, a_x, a_b, a_z, b_x, b_y, b_z, position_x, position_y, position_z, wall0_id, wall1_id, width, height, thickness, wall_id, pos_x, pos_y, size_x, size_y},
  keywords = [2]{string, bool, int, float},
  keywords = [3]{make_wall, make_wall_prim, make_door, make_bbox, make_prim,make_curved_wall},
  keywordstyle=\color{darkgray},
  keywordstyle=[2]\bfseries\color{blue},
  keywordstyle=[3]\bfseries\color{blue},
  sensitive=false,
  breaklines=true
}

\begin{document}
\newcommand{\techTerm}[1]{\emph{#1}}
\newcommand{\Aria}{\techTerm{Aria}}
\newcommand{\AStar}{\techTerm{A*}}
\newcommand{\Blender}{\techTerm{Blender}}
\newcommand{\CGAL}{\techTerm{CGAL}}
\newcommand{\COLMAP}{\techTerm{COLMAP}}
\newcommand{\Euston}{\techTerm{METHODNAME}}
\newcommand{\GeometryNodes}{\techTerm{geometry nodes}}
\newcommand{\GLB}{\techTerm{GLB}}
\newcommand{\IMU}{\techTerm{IMU}}
\newcommand{\LLM}{\techTerm{LLM}}
\newcommand{\LSDSLAM}{\techTerm{LSD-SLAM}}
\newcommand{\PyBullet}{\techTerm{PyBullet}}
\newcommand{\RGB}{\techTerm{RGB}}
\newcommand{\SLAM}{\techTerm{SLAM}}
\newcommand{\SfM}{\techTerm{SfM}}
\newcommand{\Trimesh}{\techTerm{Trimesh}}

\newcommand{\METHOD}{\texttt{SceneScript}}
\newcommand{\DatasetName}{\techTerm{Aria Synthetic Environments}}
\newcommand{\DatasetNameShort}{\techTerm{ASE}}

\definecolor{table_bbox_color}{rgb}{0.788, 0.742, 0.657}
\definecolor{sofa_bbox_color}{rgb}{0.152, 0.328, 0.536}
\definecolor{shelf_bbox_color}{rgb}{0.371, 0.524, 0.487}
\definecolor{chair_bbox_color}{rgb}{0.349, 0.573, 0.460}
\definecolor{bed_bbox_color}{rgb}{0.800, 0.680, 0.858}
\definecolor{floor_mat_bbox_color}{rgb}{0.364, 0.028, 0.243}
\definecolor{exercise_weight_bbox_color}{rgb}{0.417, 0.190, 0.343}
\definecolor{cutlery_bbox_color}{rgb}{0.599, 0.360, 0.998}
\definecolor{container_bbox_color}{rgb}{0.572, 0.788, 0.191}
\definecolor{clock_bbox_color}{rgb}{0.185, 0.021, 0.629}
\definecolor{cart_bbox_color}{rgb}{0.557, 0.555, 0.620}
\definecolor{vase_bbox_color}{rgb}{0.784, 0.853, 0.415}
\definecolor{tent_bbox_color}{rgb}{0.942, 0.780, 0.475}
\definecolor{flower_pot_bbox_color}{rgb}{0.555, 0.602, 0.953}
\definecolor{pillow_bbox_color}{rgb}{0.167, 0.193, 0.815}
\definecolor{mount_bbox_color}{rgb}{0.885, 0.425, 0.074}
\definecolor{lamp_bbox_color}{rgb}{0.360, 0.078, 0.494}
\definecolor{ladder_bbox_color}{rgb}{0.309, 0.204, 0.255}
\definecolor{fan_bbox_color}{rgb}{0.853, 0.804, 0.618}
\definecolor{cabinet_bbox_color}{rgb}{0.945, 0.184, 0.371}
\definecolor{jar_bbox_color}{rgb}{0.061, 0.831, 0.701}
\definecolor{picture_frame_bbox_color}{rgb}{0.623, 0.224, 0.610}
\definecolor{mirror_bbox_color}{rgb}{0.558, 0.264, 0.332}
\definecolor{electronic_device_bbox_color}{rgb}{0.534, 0.739, 0.043}
\definecolor{dresser_bbox_color}{rgb}{0.709, 0.031, 0.301}
\definecolor{clothes_rack_bbox_color}{rgb}{0.536, 0.392, 0.088}
\definecolor{battery_charger_bbox_color}{rgb}{0.043, 0.991, 0.131}
\definecolor{air_conditioner_bbox_color}{rgb}{0.241, 0.007, 0.729}
\definecolor{window_bbox_color}{rgb}{0.207, 0.942, 0.897}

\title{\METHOD: Reconstructing Scenes With An Autoregressive Structured Language Model}

\titlerunning{SceneScript}

\author{Armen Avetisyan\inst{1} \and
Christopher Xie\inst{1} \and
Henry Howard-Jenkins\inst{1} \and
Tsun-Yi Yang\inst{1} \and
Samir Aroudj\inst{1} \and
Suvam Patra\inst{1} \and
Fuyang Zhang\inst{2}\textsuperscript{$\dagger$} \and
Duncan Frost\inst{1} \and \\
Luke Holland\inst{1} \and
Campbell Orme\inst{1} \and
Jakob Engel\inst{1} \and
Edward Miller\inst{1} \and \\
Richard Newcombe\inst{1} \and
Vasileios Balntas\inst{1}}

\authorrunning{Avetisyan et al.}

\institute{
\textsuperscript{1} Meta Reality Labs
\qquad
\textsuperscript{2} Simon Fraser University \\
\href{https://projectaria.com/scenescript}{https://projectaria.com/scenescript}}

\maketitle

\vspace{-0.8em}
\begin{center}
    \centering
    \footnotesize
    \captionsetup{type=figure, font=small}
    \includegraphics[width=\textwidth]{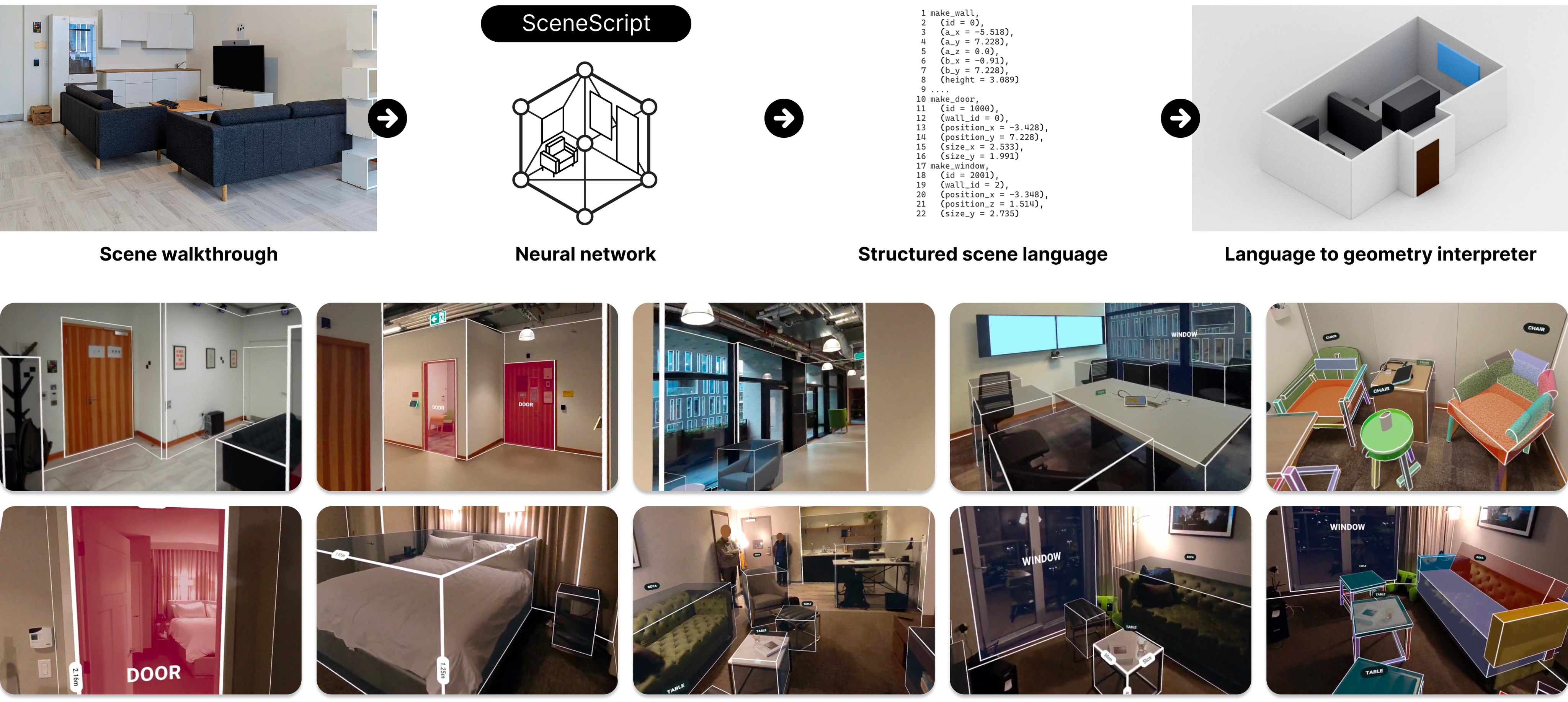}
    \captionof{figure}{\captionfont{footnotesize}(\textit{top}) Given an egocentric video of an environment, \METHOD~directly predicts a 3D scene representation consisting of structured scene language commands.
    (\textit{bottom}) Our method generalizes on diverse real scenes while being solely trained on synthetic indoor environments. 
    (\textit{last column, bottom}) A notable advantage of our method is its capacity to easily adapt the structured language to represent novel scene entities. For example, by introducing a single new command, \METHOD~can directly predict object parts jointly with the layout and bounding boxes.}
    \label{fig:teaser}
\end{center}%

\begin{abstract}

We introduce \METHOD,~a method that directly produces full scene models
as a sequence of structured language commands using an autoregressive, token-based approach. 
Our proposed scene representation is inspired by recent successes in transformers \& LLMs, and departs from more traditional methods which commonly describe scenes as meshes, voxel grids, point clouds or radiance fields. 
Our method infers the set of structured language commands directly from encoded visual data using a scene language encoder-decoder architecture. 
To train \METHOD, we generate and release a large-scale synthetic dataset  called \DatasetName~consisting of $100k$ high-quality indoor scenes, with photorealistic and ground-truth annotated renders of egocentric scene walkthroughs. Our method gives state-of-the art results in architectural layout estimation, and competitive results in 3D object detection. Lastly, we explore an advantage for \METHOD, which is the ability to readily adapt to new commands via simple additions to the structured language, which we illustrate for tasks such as coarse 3D object part reconstruction. 

\makeatletter{\renewcommand*{\@makefnmark}{}
\footnotetext{\textsuperscript{$\dagger$}Work done while the author was an intern at Meta.}\makeatother}

\end{abstract}
\vspace{-1em}

\section{Introduction}
\label{section:intro}

Scene representations play a crucial role in machine learning and computer vision applications, enabling accurate understanding of the environment. Over the years, researchers have explored various options such as meshes, voxel grids, point clouds, and implicit representations, aiming to represent complex real-world scenes with high-fidelity. Each of these exhibits distinct advantages and limitations that impact their suitability for different tasks. Meshes offer detailed geometric information but can be expensive in both computation and memory. Voxel grids provide a volumetric representation but suffer from a trade-off between resolution and memory requirements. Point clouds are efficient in representing sparse scenes but lack semantics and explicit connectivity information. Implicit representations, such as DeepSDF~\cite{park2019deepsdf} and NeRF \cite{mildenhall2021nerf}, can be infinitely precise but lack interpretability and editability. The selection of an appropriate representation directly impacts the performance and efficacy of various tasks including object recognition, scene understanding, and 3D reconstruction. In this paper, we propose a novel scene representation based on structured language commands as a more efficient and versatile solution.

Our motivation stems from the recent advancements in the field of Large Language Models (LLMs) and ``next token prediction'' autoregressive methods~\cite{openai2023gpt4}, coupled with recent works on exploring generation of sequences to represent geometric structures. For example, PolyGen \cite{nash2020polygen} demonstrated the ability to describe 3D meshes as a sequence of vertices and faces generated using transformers \cite{vaswani2017attention}. Similarly, CAD-as-Language \cite{ganin2021computer} showcased the effectiveness of generating Computer-Aided Design (CAD) primitives to represent 2D CAD sketches. 
Our main goal is to \textbf{directly infer a metrically accurate representation of a full scene} as a text-based sequence of specialized structured language commands.

Our method, denoted \METHOD, autoregressively predicts a language of hand-designed structured language commands in pure text form. This language offers several distinct advantages: 1) As pure text, it is compact and reduces memory requirements of a large scene to only a few \textbf{bytes}. 2) It is crisp and complete since the commands are designed to result in sharp and well-defined geometry (similar to scalable vector graphics). 3) It is interpretable, editable and semantically rich by design via the use of \textit{high-level parametric} commands such as \texttt{make\_door(*door\_parameters)}. 4) It can seamlessly integrate novel geometric entities by simply adding new structured commands to the language. 5) The fact that the scene representation is a series of language tokens similar to \cite{openai2023gpt4} opens up a plethora of potential new applications in the future such as ways to edit the scene, query it, or spin up chat interactions.

We mainly focus on the problems of architectural layout estimation and object detection as proxy tasks for the efficacy of our \METHOD~language as a scene representation. Architectural entities such as walls, doors, and windows are highly structured entities, making them an ideal test-bed. However, one notable drawback of language models is that they require vast amounts of data for training. Since there is no existing dataset of scene walkthroughs and their corresponding structured language commands, we publicly released \DatasetName{} (\DatasetNameShort{}),
a synthetically generated dataset of 100k unique interior scenes. For each scene, we simulate egocentric trajectories with an entire suite of sensor data from Project Aria~\cite{projectaria}. We also release multiple sources of ground truth including depth and instance segmentations. Importantly, for each rendered egocentric sequence, the architectural layout ground truth is given in our proposed \METHOD~language.

While architectural layout serves as a test-bed, we demonstrate that our method \METHOD~can easily be extended to new tasks via simple extensions to \METHOD~language while keeping both the visual input and network architecture fixed. We illustrate this on the problem of 3D object detection, which results in a method that jointly infers architectural layout and 3D oriented bounding boxes. Additionally, we demonstrate more proof-of-concept experiments that show that our method results in a significantly lower barrier to entry for new tasks including representing coarse 3D object reconstruction, curved entities, composition of entities, and entity states.

Our core contributions are:
\begin{itemize}
    \item We propose \METHOD, a method that jointly predicts architectural layout and object bounding boxes of a scene in the form of structured language commands given a video stream. 
    \item We demonstrate that \METHOD~can easily be extended 
    to completely new tasks with simple additions of commands to our structured language,
    significantly lowering the barrier to entry for new tasks.
    \item We release a large-scale synthetic dataset, named \DatasetName,
    comprized of \textbf{100k} unique high-quality 3D indoor scenes with GT,
    which will enable large scale ML training of scene understanding methods.
    \item We show that training \METHOD~on \DatasetName~leads to generalization on real scenes (
    see videos/demos on the project page).
\end{itemize}

\section{Related Works}
\label{section:related}

\begin{table*}[t]
\caption{Complete set of structured language commands designed for detailing architectural layouts and object bounding boxes. Supported data types can include \texttt{int, float, bool}. It is important to note that the language's extensibility allows for easy augmentation by introducing new commands like \texttt{make\_prim, make\_pillar}, or enhancing existing commands, such as incorporating \texttt{is\_double\_door (bool)}.}
\centering
\begin{ttfamily}
\scriptsize
\begin{tabular}{cccc}
make\_wall (\textcolor{teal}{int}) & make\_door (\textcolor{teal}{int}) & make\_window (\textcolor{teal}{int}) & make\_bbox (\textcolor{teal}{int}) 
\tabularnewline
\hline 
\hline 
\multicolumn{1}{c|}{%
\begin{tabular}{cc}
id & \textcolor{teal}{int}\tabularnewline
a\_x & \textcolor{blue}{float}\tabularnewline
a\_y & \textcolor{blue}{float}\tabularnewline
a\_z & \textcolor{blue}{float}\tabularnewline
b\_x & \textcolor{blue}{float}\tabularnewline
b\_y & \textcolor{blue}{float}\tabularnewline
b\_z & \textcolor{blue}{float}\tabularnewline
height & \textcolor{blue}{float}\tabularnewline
\end{tabular}} & \multicolumn{1}{c|}{%
\begin{tabular}{cc}
id & \textcolor{teal}{int}\tabularnewline
wall0\_id & \textcolor{teal}{int}\tabularnewline
wall1\_id & \textcolor{teal}{int}\tabularnewline
position\_x & \textcolor{blue}{float}\tabularnewline
position\_y & \textcolor{blue}{float}\tabularnewline
position\_z & \textcolor{blue}{float}\tabularnewline
width & \textcolor{blue}{float}\tabularnewline
height & \textcolor{blue}{float}\tabularnewline
\end{tabular}} & \multicolumn{1}{c|}{%
\begin{tabular}{cc}
id & \textcolor{teal}{int}\tabularnewline
wall0\_id & \textcolor{teal}{int}\tabularnewline
wall1\_id & \textcolor{teal}{int}\tabularnewline
position\_x & \textcolor{blue}{float}\tabularnewline
position\_y & \textcolor{blue}{float}\tabularnewline
position\_z & \textcolor{blue}{float}\tabularnewline
width & \textcolor{blue}{float}\tabularnewline
height & \textcolor{blue}{float}\tabularnewline
\end{tabular}} & \multicolumn{1}{c}{%
\begin{tabular}{cc}
id & \textcolor{teal}{int}\tabularnewline
class & \textcolor{teal}{int}\tabularnewline
position\_x & \textcolor{blue}{float}\tabularnewline
position\_y & \textcolor{blue}{float}\tabularnewline
position\_z & \textcolor{blue}{float}\tabularnewline
angle\_z & \textcolor{blue}{float}\tabularnewline
scale\_x & \textcolor{blue}{float}\tabularnewline
scale\_y & \textcolor{blue}{float}\tabularnewline
scale\_z & \textcolor{blue}{float}\tabularnewline
\end{tabular}} \tabularnewline


\end{tabular}

\end{ttfamily}
\label{table:commands_and_parameters}
\end{table*}

\subsection{Layout Estimation}
Layout estimation is an active research area, aiming to infer architectural elements. Scan2BIM~\cite{murali2017indoor} proposes heuristics for wall detection to produce 2D floorplans.
Ochmann et al.~\cite{ochmann2019automatic} formulate layout inference as an integer linear program using constraints on detected walls. 
Shortest path algorithms around birds-eye view (BEV) free space~\cite{cabral2014piecewise} and wall plus room instance segmentation~\cite{chen2019floor} have also been explored.

Furukawa et al.~\cite{furukawa2009reconstructing} utilize a \textit{Manhattan world}-based multi-view stereo algorithm~\cite{furukawa2009manhattan} to merge axis-aligned depth maps into a full 3D mesh of building interiors.
RoomNet~\cite{lee2017roomnet} predicts layout keypoints while assuming that a fixed set of Manhattan layouts can occur in a single image of a room. 
LayoutNet~\cite{zou2018layoutnet} improves on this by predicting keypoints and optimising the Manhattan room layout inferred from them.
Similarly, AtlantaNet~\cite{pintore2020atlantanet} predicts a BEV floor or ceiling shape and approximates the shape contour with a polygon resulting in an \textit{Atlanta world} prior. 
SceneCAD~\cite{avetisyan2020scenecad} uses a graph neural network to predict object-object and object-layout relationships to refine its layout prediction.

Our approach stands out by requiring neither heuristics nor explicitly defined prior knowledge about architectural scene structure.
In fact, our method demonstrates geometric understanding of the scene which emerges despite learning to predict the GT scene language as a sequence of text tokens.

\subsection{Geometric Sequence Modelling}

Recent works have explored transformers for generating objects as text-based sequences.
PolyGen~\cite{nash2020polygen} models 3D meshes as a sequence of vertices and faces. 
CAD-as-Language~\cite{ganin2021computer} represents 2D CAD sketches as a sequence of triplets in protobuf representation, followed by a sequence of constraints. 
Both SketchGen~\cite{para2021sketchgen} and SkexGen~\cite{xu2022skexgen} use transformers to generate sketches.
DeepSVG~\cite{carlier2020deepsvg} learns a transformer-based variational autoencoder (VAE) that is capable of generating and interpolating through 2D vector graphics images. 
DeepCAD~\cite{wu2021deepcad} proposes a low-level language and architecture  similarly to DeepSVG, but applies it to 3D CAD models instead of 2D vector graphics.
Our approach stands out by utilising \textit{high-level} commands, offering interpretability and semantic richness. Additionally, while low-level commands can represent arbitrarily complex geometries, they lead to prohibitively longer sequences when representing a full scene.

The closest work to ours is Pix2Seq~\cite{chen2021pix2seq}. Pix2Seq proposes a similar architecture to ours but experiments only with 2D object detection, thus requiring domain-specific augmentation strategies. Another closely related work is Point2Seq~\cite{xue2022point2seq} that trains a recurrent network for autoregressively regressing continuous 3D bounding box parameters. Interestingly, they find the autoregressive ordering of parameters outperforms current standards for object detection architectures, including anchors~\cite{girshick2015fast} and centers~\cite{yin2021center}.

\section{\METHOD~Structured Language Commands}
\label{section:ssl}

\begin{figure*}[t]
\includegraphics[width=\textwidth]{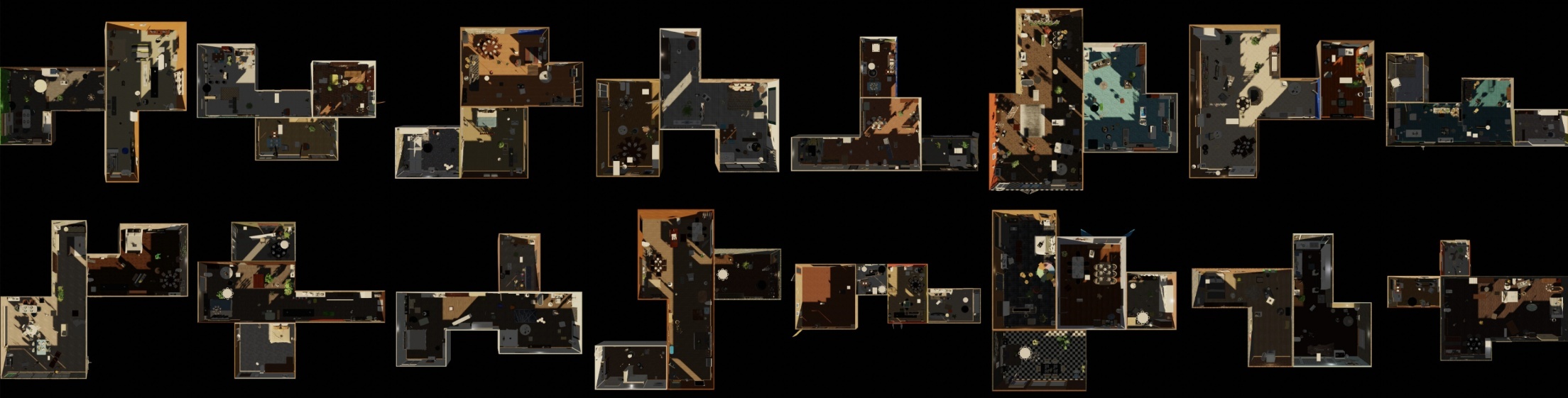}
\includegraphics[width=0.245\textwidth]{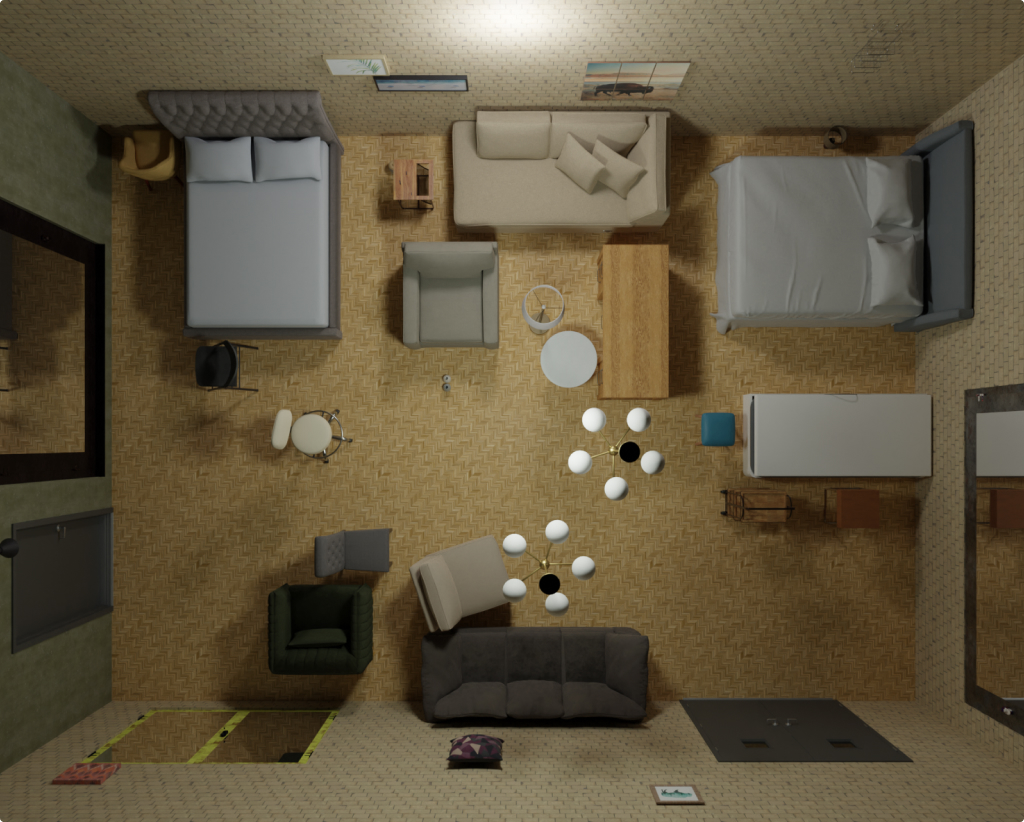} 
\includegraphics[width=0.245\textwidth]{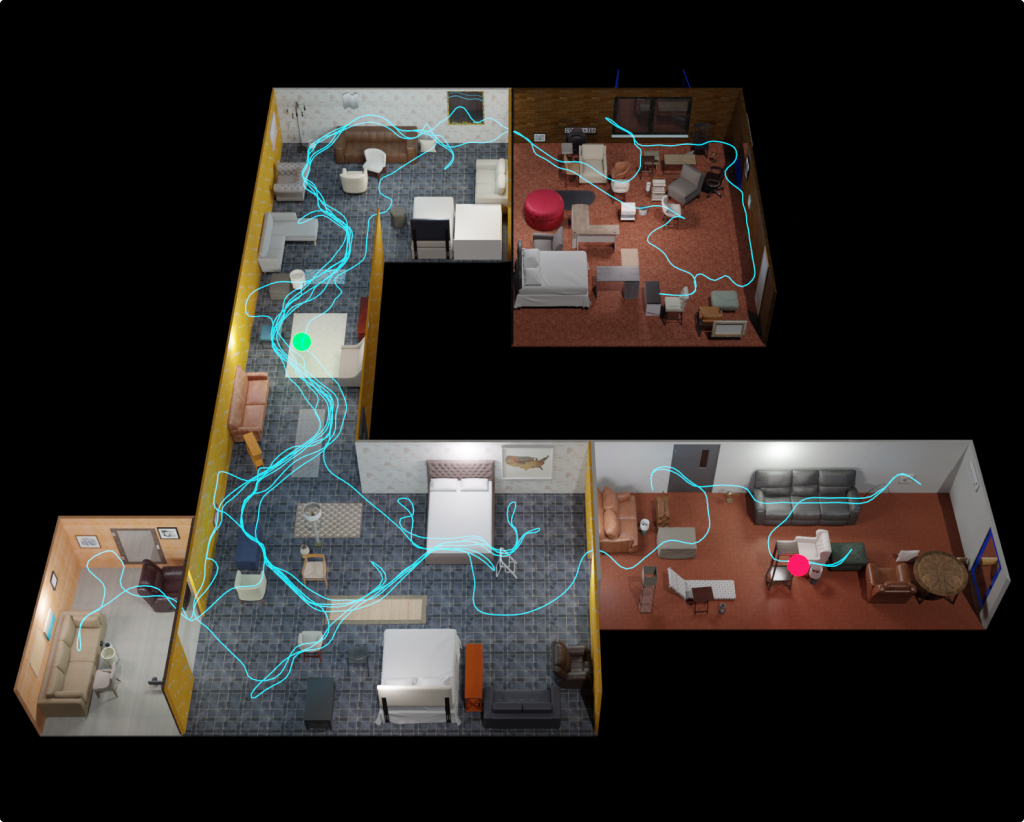} 
\includegraphics[width=0.245\textwidth]{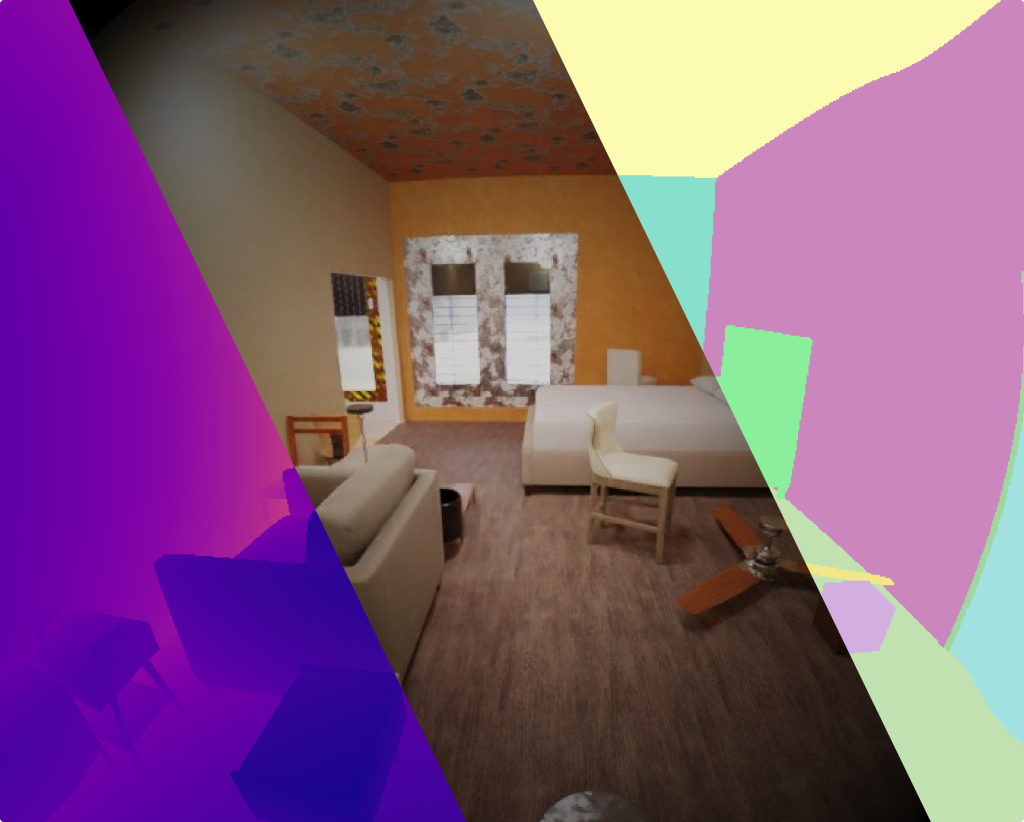} 
\includegraphics[width=0.245\textwidth]{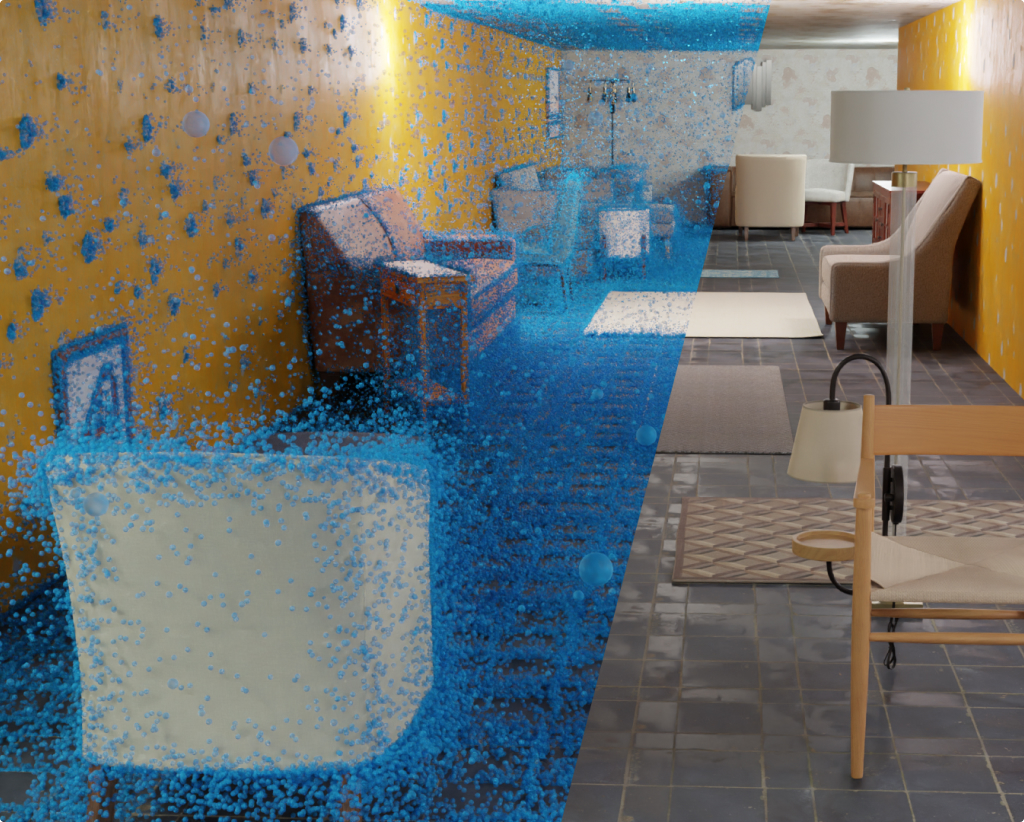}
\caption{\DatasetName:
(top) Random samples of generated scenes showing diversity of layouts, lights and object placements. (bottom - left to right) A top down view of a scene filled with objects, a simulated trajectory (blue path), renderings of depth, RGB, and object instances, and lastly a scene pointcloud.}
\label{fig:dataset_teaser}
\end{figure*}

We first describe our structured language commands
that define a full scene representation including both layout and objects.
After this, we introduce our corresponding
large scale training dataset: \DatasetName{}.

\subsection{Commands and Parameters}
\label{subsec:ssl_commands}

We begin with a parameterization that captures the most common layout elements.
For this purpose we use three commands:
\texttt{make\_wall}, \texttt{make\_door}, and \texttt{make\_window}.
Each command comes with a set of parameters that results in well-defined geometry.
For example,
the full set of parameters for \texttt{make\_wall} specifies a gravity-aligned 2D plane,
while \texttt{make\_door} and \texttt{make\_window} specify box-based cutouts from walls.
It is worth noting that this parametrization is arbitrary,
and is only made in the context of presenting a proof-of-concept \METHOD~system.
There are infinitely many parametrization schemes,
in this work we opt for one that prioritizes ease of use and research iteration speed.

In addition to representing these three major layout entities, we aim to jointly infer objects as oriented bounding boxes. Thus, we introduce a fourth command:
\begin{lstlisting}[language=StructuredLanguage]
make_bbox: id, class, position_x, position_y, position_z, angle_z, scale_x, scale_y, scale_z
\end{lstlisting}
This simple parametrization represents an oriented 3D bounding box that is assumed to be aligned with gravity (assuming it points in the $-z$ direction). 
A summary of these commands and their respective parameters are shown in Table~\ref{table:commands_and_parameters}.

While we have described just four commands to capture structure and objects in an indoor environment, importantly,
this text-based parametrization can readily be extended 
geometrically and/or even semantically
to include states or other functional aspects.
For example,
changing the mentioned \texttt{make\_door} command
to include parameters 
such as \texttt{open\_degree}
allows the language to represent door states.
In Section~\ref{section:extendability},
we demonstrate how such extensions to the language allows for \METHOD~to readily adapt to new tasks including coarse 3D object reconstruction.

\subsection{Scene Definition}

A single scene can be described as a sequence of our proposed structured language commands. The sequence requires no specific ordering, and can be of arbitrary length.
From this definition, the 3D scene can easily be obtained by parsing the commands through a simple custom interpreter.

\subsection{Training Dataset}
\label{section:data}

To enable practical indoor scene reconstruction
based on our structured language commands
(Section~\ref{subsec:ssl_commands}),
we publicly released a new dataset called \DatasetName.
It consists of a large number of training pairs
of egocentric scene walkthroughs
linked with corresponding ground truth command sequences.

Since transformers require vast amounts of data,
we generated $100k$ synthetic scenes,
which is in comparison infeasible for real-world data.
Each synthetic scene comes with a floor plan model, 
a corresponding complete 3D scene
 as well as a simulated agent trajectory and
a photo-realistic rendering of this trajectory.
Fig. \ref{fig:dataset_teaser} illustrates
the basics of \DatasetName{}.
For brevity, we refer the reader
to 
Appendix~\ref{sec:input_data}
for further details.

\section{\METHOD~Network Architecture}
\label{section:method}

\begin{figure*}[t]
    \centering
    \includegraphics[width=\textwidth]{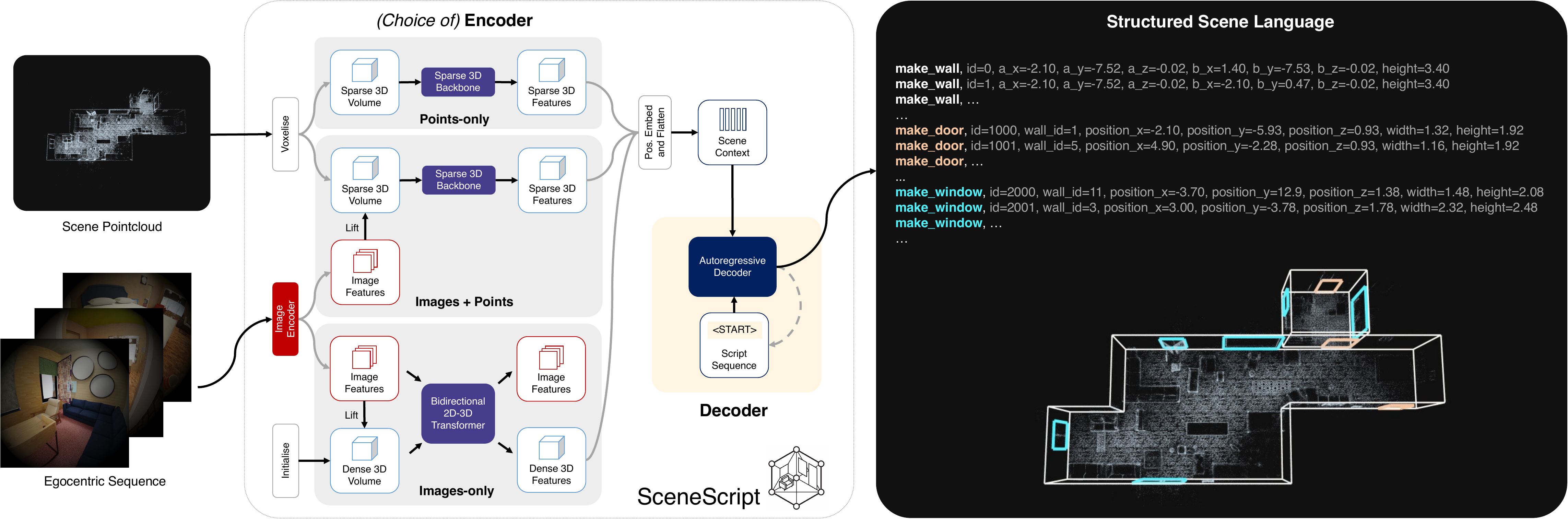}\\[-3mm]
    \caption{\METHOD~core pipeline overview. Raw images \& pointcloud data are encoded into a latent code, which is then autoregressively decoded into a sequence of commands that describe the scene. 
    Visualizations are shown using a customly built interpreter. Note that for the results in this paper, the the point clouds are computed from the images using Aria MPS \cite{aria_white_paper} -- i.e. are not using a dedicated RGB-D / Lidar sensor.}
    \label{fig:main_pipeline}
\end{figure*}

Our pipeline is a simple encoder-decoder architecture that consumes a video sequence and returns \METHOD~language in a tokenized format. Figure~\ref{fig:main_pipeline} illustrates a high-level overview of our method.

We examine three encoder variants: a pointcloud encoder, a posed image set encoder, and a combined encoder. The decoder remains the same in all cases.

\subsection{Input Modalities and Encoders}
\label{sec:encoders}
The encoder computes a latent scene code
in the form of a 1D sequence
from video walkthrough of the scene.
The decoder is designed
to consume these 1D sequences as input.
This enables the integration of various input modalities
within a unified framework.
As a preliminary,
for each scene we assume access to
a set of $M$ posed camera images
$\{\mathbf{I_1},...,\mathbf{I_M}\}$, 
e.g., \SLAM{} output. 

\subsubsection{Point Clouds.}
\label{sec:encoder-points-only}

A point cloud 
$\mathbf{P} = \{\mathbf{p_1}, ..., \mathbf{p_N}\}$
consists of $N$ points,
where $\mathbf{p_i}$ is a 3D point coordinate. It can come from passive images using \SLAM{} or \SfM{}, or RGB-D / Lidar sensors.

Specifically, we use the Semi-dense Pointclouds from Project Aria's Machine Perception Services~\cite{aria_white_paper},
that are obtained from a visual-inertial SLAM system using Aria's monochrome cameras and IMUs. We discretize the point cloud to 5cm resolution, then employ a sparse 3D convolution library~\cite{tang2022torchsparse,tang2020searching} to generate pooled features. The encoder $\mathcal{E}_{geo}$ applies a series of down convolutions, resulting in a reduction of the number of points in the lowest level. 

\begin{equation}
\mathbf{F}_{geo} = \mathcal{E}_{geo}(\mathbf{P}), \quad \mathbf{P} \in \mathbb{R}^{N \times 3}, \mathbf{F}_{geo} \in \mathbb{R}^{K \times 512}
\end{equation}
where $K \ll N$. $\mathbf{F}_{geo}$ is a condensed latent representation of the point cloud that contains the necessary scene context. For later use in the transformer decoder, we treat $\mathbf{F}_{geo}$ as a sequence of feature vectors where the entries $\mathbf{f_i},\; i \in {1 ... K}$ are sorted lexicographically according to the coordinate of the active site $\mathbf{c_i},\; i \in {1 ... K}$. To incorporate positional encoding, we append the coordinates of the active sites to their respective feature vectors $\mathbf{f}_i \leftarrow \texttt{cat}(\mathbf{f_i}, \mathbf{c_i})$.

\subsubsection{Point Clouds with Lifted Features.}

We additionally explore augmenting the point cloud with image features. From the original egocentric sequence and associated trajectory, we sample a set of $M$ keyframes, $\mathbf{I}_i$ where $i \in {1 ... M}$, and compute a set of image features $\mathbf{F}_i$ for each. We then project each point into the set of keyframe cameras and retrieve the feature vector (output by a CNN) at the pixel location:
\begin{equation}
    \mathbf{f}_{ip} = F_i(\pi(\mathbf{p})) \qquad \mathbf{p}\in\mathbf{P}, i \in {1 ... M},
\end{equation}
where $\pi(\cdot)$ represents the projection function of a 3D point into the camera. If $\pi(\mathbf{p})$ falls outside the image bounds, no feature is retrieved. We combine the set of lifted features for a point through an average, resulting in a single feature vector for each point: 
$\mathbf{f}_p = 1/M\sum_{i=1}^{M}{\mathbf{f}_{ip}}$.

We form our lifted-feature point cloud, $\mathbf{P'} = \{\mathbf{p'_1}, ..., \mathbf{p'_M}\}$, by  concatenating each point's lifted feature with the original XYZ location: $\mathbf{p}' = \texttt{cat}(\mathbf{f}_p,\mathbf{p})$. $\mathbf{P'}$ is then encoded into a context sequence using sparse 3D convolutions, with only the input feature channels adjusted to match the new point feature dimension.

\subsubsection{End-to-end Encoding of Posed Views.}
In order to encode the egocentric sequence more directly without a pre-computed point cloud, we adopt a 2D~$\leftrightarrow$~3D bidirectional transformer encoder following the form defined in RayTran~\cite{tyszkiewicz2022raytran}. 

In this formulation, we initialize a volume representation of the scene as a dense voxel grid of features, $\mathbf{V}$, that coincides with the scene geometry. In turn, we sample a subset of $M$ keyframes, $\mathbf{I}_i$ where $i \in {1 ... M}$, from the full stream of posed images. And for each of these keyframes we compute image features from a CNN, $\mathbf{F}_i$. 
Repeated layers of bidirectional attention enable the image and voxel grid features to be refined iteratively in successive transformer blocks through the aggregation of view-point and global-scene information. As in RayTran~\cite{tyszkiewicz2022raytran}, the interaction between the two representations is guided by the image-formation process by leveraging known camera parameters and poses. Attention in these transformer blocks is restricted by patch-voxel ray intersections, where each image patch attends to the voxels it observes and each voxel location attends to all the patches that observe it. The resulting voxel grid of features is flattened, concatenated with an encoded represention of its XYZ location, and passed to the decoder.

\subsection{Language Decoder}
\label{section:decoder}
We utilize a transformer decoder~\cite{vaswani2017attention} to decode the scene latent code into a sequence of structured language commands.
The sequence of tokens passes through an embedding layer, followed by a positional encoding layer. 
Together with the encoded scene code (Section~\ref{sec:encoders}), the embedded tokens are passed into the several transformer decoder layers where a causal attention mask is used to ensure autoregressive generation. More implementation details can be found in 
Appendix~\ref{app:decoder}.

\subsection{Language Tokenization}
\label{section:tokenization}
We refer to the serialization the structured language into a sequence of tokens as \textbf{tokenization}. The goal is to construct a bijective mapping between a sequence of structured language commands (Section~\ref{section:ssl}) and a sequence of integer tokens that can be predicted by the transformer decoder architecture. 
We utilize the following schema:
\begin{align*}
    &[ \paramstyle{col4}{START}, \paramstyle{col1}{PART}, \paramstyle{col2}{CMD},  \paramstyle{col2}{PARAM\_1},
    \paramstyle{col2}{PARAM\_2}, \ldots, \paramstyle{col2}{PARAM\_N}, \paramstyle{col1}{PART}, \ldots, \paramstyle{col4}{STOP} ] 
\end{align*}

For example, a sample sequence for a \texttt{make\_door} command may look like:
\begin{align*}
    &[ \paramstyle{col4}{START}, \paramstyle{col1}{PART}, \paramstyle{col2}{MAKE\_DOOR}, \paramstyle{col2}{POSITION\_X}, \paramstyle{col2}{POSITION\_Y}, \paramstyle{col2}{POSITION\_Z},\\
    &\paramstyle{col2}{WALL0\_IDX},  \paramstyle{col2}{WALL1\_IDX}, \paramstyle{col2}{WIDTH}, \paramstyle{col2}{HEIGHT},\paramstyle{col1}{PART}, \ldots, \paramstyle{col4}{STOP} ] 
\end{align*}
%
This schema enables 1D packing of tokens without requiring fixed-size slots or padding like other sequence modelling methods such as \cite{wu2021deepcad}. Additionally, it does not impose any limitations on the number or the hierarchy of sub-sequences, as they are flexibly separated by a \paramstyle{col1}{PART} token. This allows for arbitrarily complex scene representations.

The tokenized sequence is discretized into integers at a 5cm resolution, then translated into a sequence of embeddings via learnable lookup table. Note that by designing the \METHOD~language, we also design the tokenization. This tokenization scheme is notably different from standard NLP tokenization, which involves Byte-Pair Encodings (BPE) \cite{openai2023gpt4}.

\section{Results}
\label{section:experiments}

In this section, we introduce the metrics we define to measure performance, and we discuss some qualitative and quantitative results that give insights to our proposed \METHOD~method.

\subsection{Metrics}
To evaluate accuracy of the layout estimation, we make use of geometric metrics applied between the ground-truth room layout and our predicted \METHOD~language. To do so, we define an \textit{entity distance}, $d_E$, between a pair of entities of the same class. Each entity, $E$, is represented as a 3D plane segment comprising of 4 corners $\{c_1, c_2, c_3, c_4\}$. The distance between two entities, $E$ and $E'$ is computed as the maximum Euclidean distance between each corner and its counterpart assigned via Hungarian mathcing, i.e.: $d_E(E, E') = \text{max}\{||c_i - c'_{\pi(i)}|| : i = 1, ..., 4\}$, where $\pi(i)$ is the permutation also found via Hungarian matching. 

We threshold $d_E$ to define the success criteria for the predicted entities. We compute the following metrics: 
\begin{itemize}
    \item F1 Score @ \textit{threshold} -- the F1 score of the set of predictions is computed at a single $d_E$ threshold.
    \item Average F1 Score -- the F1 score is computed across a range of entity distance thresholds and averaged.
\end{itemize}

The scores are computed for each class independently and averaged to overall score. In addition, scores are computed for each scene and averaged across the dataset. We use the following range of thresholds (cm) for the average F1 scores: $T = \{1, 2, ..., 9, 10, 15, 25, 30, 50, 75, 100\}$.

\begin{figure*}[t]
    \centering
    \includegraphics[width=\linewidth]{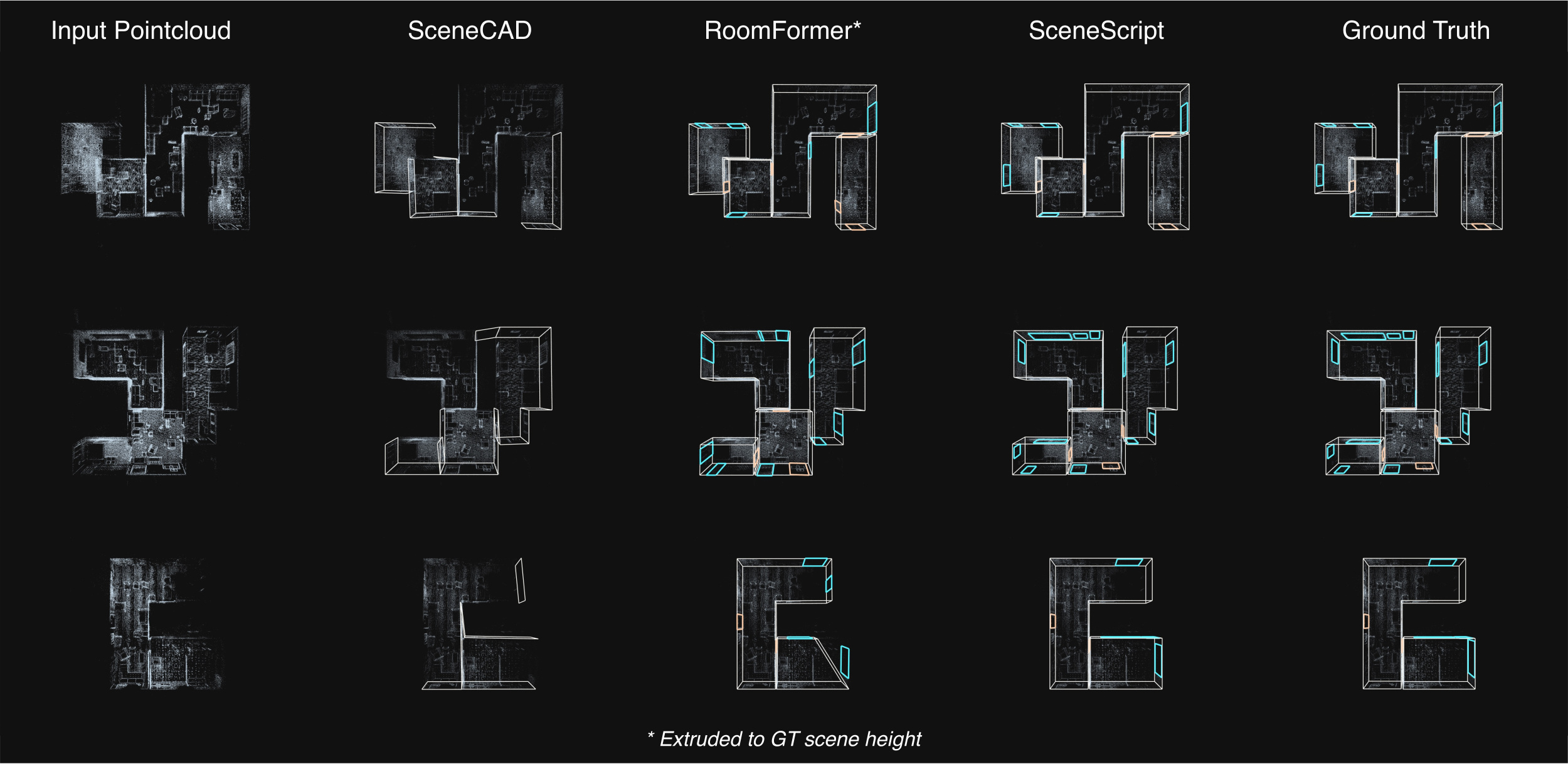}
    \caption{Qualitative samples between our model and SOTA methods on \DatasetName's test set. Hierarchical methods like SceneCAD suffer from error cascading which leads to missing elements in the edge prediction module. RoomFormer (a 2D method extruded to 3D) primarily suffers from lightly captured scene regions which leave a unnoticeable signal in the density map. }
    \label{fig:qualitative_layout}
\end{figure*}

\subsection{Layout Estimation}

\begin{table*}[t]
    \centering
    \caption{\textbf{Layout Estimation on \DatasetName{}} Quantitative comparison between our method and related recent work. 
    }
    \begin{tabular}{c|cccc|cccc}
         & \multicolumn{4}{c|}{F1 @5cm} & \multicolumn{4}{c}{Avg F1}\\
        Method & mean & wall & door & window & mean & wall & door & window \\
        \hline
        \hline
        SceneCAD '20~\cite{avetisyan2020scenecad}           & - & 0.048   & -  & -    & - & 0.275 & - & - \\
        RoomFormer '23~\cite{yue2023connecting}             & 0.139 & 0.159 & 0.148 & 0.110     & 0.464 & 0.505 & 0.481 & 0.407 \\
        \hline
        Ours (Point cloud)     & 0.848 & 0.930 & 0.922 & 0.692     & 0.784 & 0.816 & 0.811 & 0.724 \\
        Ours (Lifted features)  & \textbf{0.903} & \textbf{0.943} & \textbf{0.959} & \textbf{0.806}     & \textbf{0.801} & \textbf{0.818} & \textbf{0.822} & \textbf{0.764}\\
        Ours (Image-only)       & 0.661 & 0.687 & 0.798 & 0.497     & 0.719 & 0.727 & 0.772 & 0.658 \\

    \end{tabular}
    \label{tab:main_results_table}
\end{table*}

We perform scene layout estimation with the the three encoder variants of \METHOD: a baseline model with sparse3d convolution pointcloud encoder, an RGB RayTran-based feature volume encoder~\cite{tyszkiewicz2022raytran}, and our proposed lifted feature point encoder. The same transformer decoder is being used in all three scenarios. 

To provide comparison to existing works, we include results from two baseline methods, namely SceneCAD~\cite{avetisyan2020scenecad} and the recent  RoomFormer~\cite{yue2023connecting}. For these experiments SceneCAD and RoomFormer were both trained on \DatasetName. 
Note that SceneCAD only predicts walls.

Table \ref{tab:main_results_table} shows the main results for our F1-based metrics on \DatasetName{}. \METHOD~exhibits a substantial performance advantage over SOTA layout estimation baselines across multiple metrics. 
Both baseline methods encounter a significant decline in accuracy when dealing with finer details. See Figure~\ref{fig:qualitative_layout} for qualitative comparisons between our method and baseline methods. 


\subsubsection{Encoder Ablation.} 
The results demonstrate that \METHOD~is robust to the encoding strategy chosen to encode the egocentric capture. It is able to infer highly accurate scene layouts in all configurations tested, and in each case \METHOD~outperforms the included baselines by a significant margin. 

Relative comparison of the encoder strategies reveals that leveraging the pointclouds from a highly specialized and well-tuned system is still able to offer advantages of an, almost, entirely learned approach such as RayTran~\cite{tyszkiewicz2022raytran}. A light extension in the form of lifted image features can widen this gap even further. In particular, we observe that the discrepancy between the encoders becomes particularly apparent as the complexity of the scene increases in the form of increased room count -- more details are included in 
Appendix~\ref{app:num_rooms}.
In Appendix~\ref{app:num_rooms}, we also show a quantitative evaluation 
of per-entity error distances, which aids in further attribution of relative performance gains between the encoding methods.

\begin{table}[t]
\centering
    \caption{\textbf{3D Object Detection} Performance comparison against state-of-the-art methods on an 3D object detection task trained and evaluated by F1-score at 0.25 and 0.5 IoU thresholds. By simply adding a \texttt{make\_bbox} command \METHOD{} can achieve competitive object detection results.
    }
    \begin{subtable}{0.48\linewidth}
    \centering
        \caption{\DatasetName}
        \resizebox{\textwidth}{!}{%
        \begin{tabular}{c|c|cc}
            & & \multicolumn{2}{c}{F1} \\
             Method & Input & @.25 IoU & @.50 IoU \\
            \hline \hline
            3DETR '21~\cite{misra2021end} & Points & 0.201 & 0.078 \\
            Cube R-CNN '23~\cite{brazil2023omni3d} & RGB &  0.394 & 0.228 \\
            ImVoxelNet '22~\cite{rukhovich2022imvoxelnet} & RGB & 0.584 & 0.516\\
            Ours & Points & \textbf{0.620} & \textbf{0.577} \\
        \end{tabular}
        }
        \label{tab:obj_det}
    \end{subtable}
    \begin{subtable}{0.48\linewidth}
    \centering
        \caption{ScanNet~\cite{dai2017scannet}}
        \resizebox{\textwidth}{!}{%
        \begin{tabular}{c|c|cc}
          & & \multicolumn{2}{c}{F1} \\
         Method & Input & @.25 IoU & @.50 IoU \\
         \hline \hline
         3DETR '21~\cite{misra2021end} & Points & 0.480 & 0.349\\
         3DETR-m '21~\cite{misra2021end} & Points & 0.536 & 0.407 \\
         SoftGroup '22~\cite{vu2022softgroup} & RGB Points & \textbf{0.622} & \textbf{0.573} \\
         Ours & RGB Points & 0.506 & 0.406 \\
        \end{tabular}
        }
        \label{tab:results_object_detection_scannet}
    \end{subtable}
\end{table}

\subsection{Object Detection}
In this section, we perform evaluation of \METHOD{} for object detection on both \DatasetName{} and ScanNet~\cite{dai2017scannet}. For comparison, we include recent and state-of-the-art baseline methods, namely Cube-RCNN~\cite{brazil2023omni3d}, ImVoxelNet~\cite{rukhovich2022imvoxelnet}, 3DETR~\cite{misra2021end}, and SoftGroup~\cite{vu2022softgroup}. 

Worth noting is that \METHOD{} does not predict a confidence score for each \texttt{make\_bbox} command. This means that fair computation of the conventional mAP metric for this task is not possible, as detections cannot be ranked across scenes. Among other issues, this results in a metric that varies with the order in which scenes are evaluated. We therefore report F1-score-based metrics, which do not exhibit this order variance. Further discussion of this, and mAP numbers for the baselines for reference, can be found in 
Appendix~\ref{app:baseline_map}.



In Table~\ref{tab:obj_det}, all methods were trained on \DatasetName. 
3DETR performs poorly due to its encoder being designed for relatively clean datasets such as ScanNet~\cite{dai2017scannet}, while semi-dense point clouds from \DatasetName{} exhibit more noise and less uniformity (see Figure~\ref{fig:main_pipeline} for an example).
Cube R-CNN and ImVoxelNet both operate on RGB images, and detections are tracked for the entire sequence via a tracker~\cite{brazil2023omni3d} to provide competitive performance. 
In Table~\ref{tab:results_object_detection_scannet}, our method provides similar performance to both 3DETR and SoftGroup.

Through the addition of the \texttt{make\_bbox} command, \METHOD~demonstrates object detection performance on par with SOTA baselines on multiple object detection datasets. 
This result illustrates the extensibility of a token-based approach, and that our proposed \METHOD~language representation does not suffer compared to specialised object detection network architectures.

\section{Extending the \METHOD~Structured Language}
\label{section:extendability}

\begin{figure}[t]
    \centering
    \includegraphics[width=\textwidth]{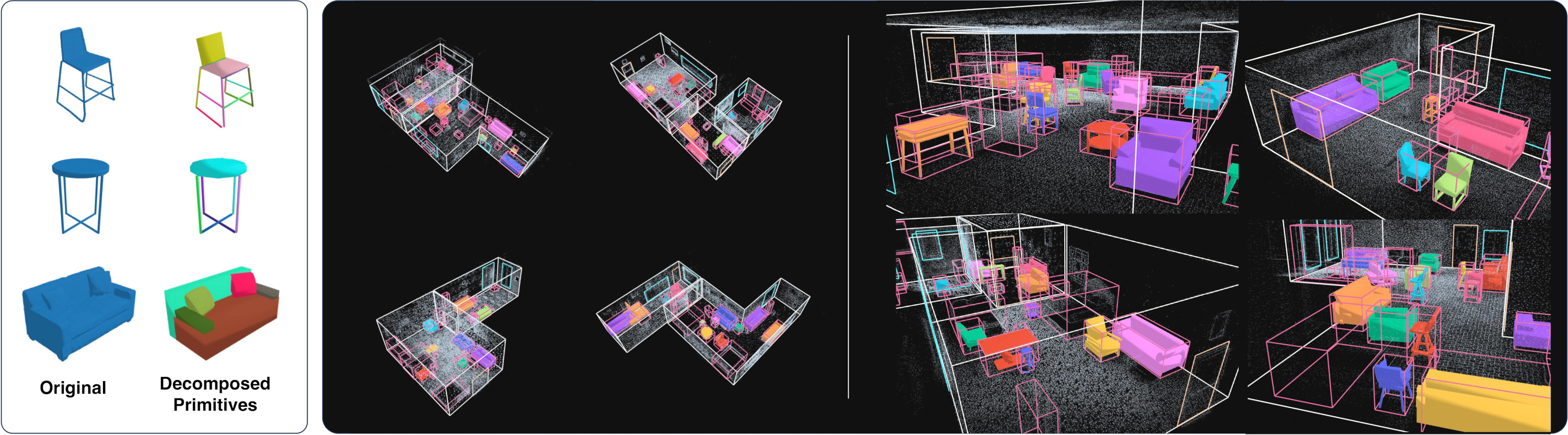}
    \caption{Example scene reconstructions on scenes from \DatasetName{}.
        (left) Visualisation of the decomposed meshes used to create \texttt{make\_prim} training pairs.
        (right) Views of full scene predictions, as well as close ups highlighting the fidelity of object reconstruction through the prediction volumetric primitives enabled by \texttt{make\_prim}.}
    \label{fig:commandVisExamples1}
\end{figure}

A key advantage offered by \METHOD's structured language prediction paradigm is that the expressiveness of its reconstruction can be tailored without requiring a change to the method. Up to now, we have focussed on showcasing the efficacy of \METHOD~for representing simple layout elements and objects as bounding boxes. In this section, we showcase this characteristic by increasing the fidelity of our scene representation by introducing coarse 3D object reconstruction.

\subsection{Objects as Volumetric Primitives}
We turn to a language based on volumetric primitives, motivated by works such as~\cite{tulsiani2017learning,yang2021unsupervised}. 
Using simple primitives such as cuboids and extruded cylinders, enables us to coarsely represent arbitrary object categories while maintaining object semantics (e.g. tabletops can be represented by a single cuboid). 
Thus, this language can describe many object categories simultaneously.

This representation requires only one additional command over the layout and box commands already discussed previously, namely:
\begin{lstlisting}[language=StructuredLanguage]
make_prim: bbox_id, prim_num, class, center_x, center_y, center_z, angle_x, angle_y, angle_z, scale_x, scale_y, scale_z
\end{lstlisting}
This command and its parameters describe a volumetric primitive (cuboid or extruded cylinder) via its 3D center, 3D rotation, and 3D scale. The \texttt{prim\_num} parameter can be associated with semantics, e.g. tabletops of different tables typically have the same \texttt{prim\_num}.

\begin{figure*}[t]
    \centering
    \includegraphics[width=\textwidth]{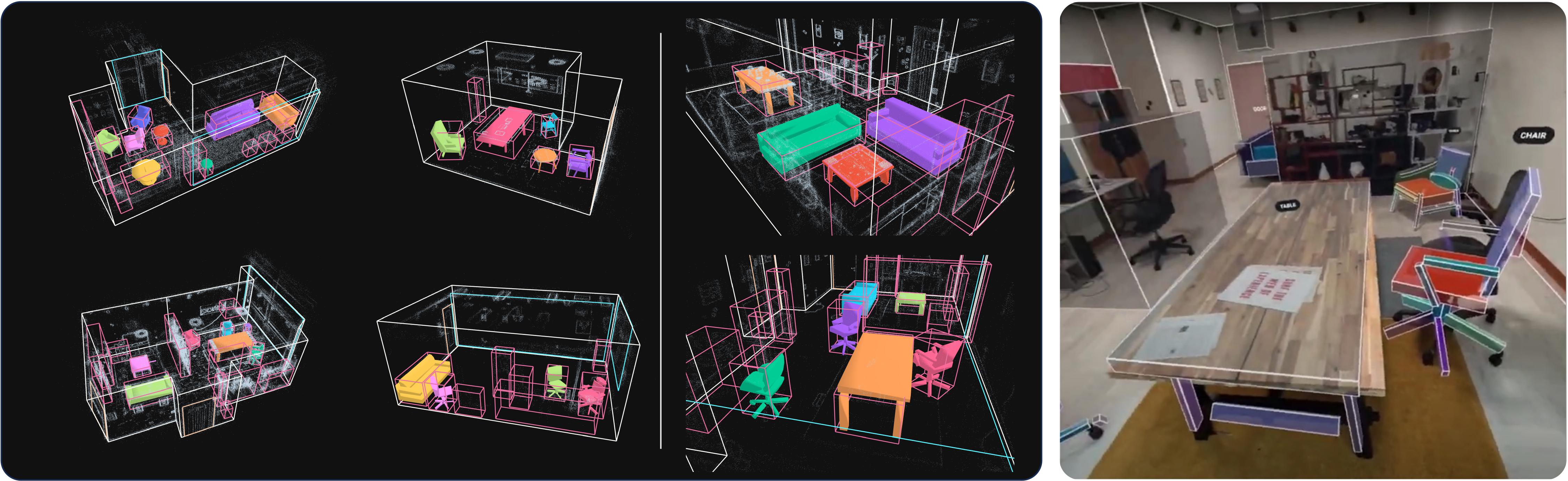}
    \caption{Example scene reconstructions on \textbf{real scenes} with the addition of the \texttt{make\_prim} command. Note that \METHOD{} was trained only on synthetic data.}
    \label{fig:extensions}
\end{figure*}

\subsubsection{Dataset.}
To obtain ground truth \texttt{make\_prim} commands that align with the objects in \DatasetName, we first run an extension of Yang~\etal~\cite{yang2021unsupervised} to obtain cuboid and extruded cylinder primitives of a database of 3D CAD models (ABO~\cite{collins2022abo}, which was used to populate \DatasetName). See Figure~\ref{fig:commandVisExamples1} (left) for example decompositions. For this proof-of-concept experiment, we use three categories: \textit{table}, \textit{chair}, and \textit{sofa}. We then convert these decomposed primitives into \texttt{make\_prim} commands that are aligned with the objects in the dataset, which results in training pairs.

\subsubsection{Results.} 
We show qualitative results on \DatasetName{} in Figure~\ref{fig:commandVisExamples1}.
In Figure~\ref{fig:extensions},
we show inferences in a few real-world environments
despite only having trained on our simulated dataset.
These results demonstrate that \METHOD{}'s general purpose architecture is capable of coarsely reconstructing objects of multiple categories through addition of a new command.

\subsection{Further Extensions}
In this section, we have explored just one extension of \METHOD's structured scene language in order to demonstrate its flexibility. The result was greatly increased expressiveness of the scene reconstruction. 
In Appendix~\ref{app:extensions},
we include additional explorations that can further increase the fidelity and accuracy of \METHOD's reconstructions. These explorations include reconstructing curved walls, inferring object states (such as door opening angles), as well as direct prediction of parametric models using object models deployed commonly by tech artists (e.g., blender geometry nodes) for object reconstruction.

\section{Interactive Scene Reconstruction}

Scene reconstruction often occurs offline, on pre-recorded walkthroughs of an environment, or derivatives of them, such as a point cloud. 
However, we take advantage of \METHOD's inference occurring at an interactive rate, which takes between 2-5s depending on the size of the environment, by implementing streaming of \METHOD's live reconstructions into a modern VR headset. This enables an interactive experience where the user can see the reconstruction overlaid onto the environment they are exploring in real-time. See the video recording 
on the website
for examples.

With this the user can actively refine the results, for example through more thorough exploration of areas that may have been missed. Visualisations of this interface are included in the bottom half for Figure~\ref{fig:teaser}.

\section{Limitations and Future Work}
\label{section:limitations_future_work}

\METHOD~exhibits certain limitations that should be acknowledged. First, the structured language commands are manually defined, which requires human intervention at this stage. Secondly, due to the higher-level nature of our commands, it can be challenging to capture fine-grained geometric details with extremely high precision (i.e. mm). As a consequence, the resulting reconstructions based on this representation tend to lead to simpler and coarser geometries, potentially missing intricate nuances at the very high detail level. 
These limitations potentially highlight areas for future research and optimization, aiming to automate the command definition process and explore techniques to enhance the representation's ability to capture intricate geometric details accurately. However, we believe that the ability to built scene representations that are based on structured language commands will be a key part in complex and efficient methods of scene reconstruction in the future, enabling them to be used in conjunction with general purpose LLMs.

\section{Conclusion}
\label{section:conclusion}

We introduced \METHOD, a novel reconstruction method that is based on a tokenized scene representation. \METHOD{} autoregressively predicts a sequence of structured scene language commands given a video stream of an indoor space. This tokenized scene representation is compact, editable and intepretable. In addition, we showed that a strength of \METHOD{} is the ability to extend to arbitrary elements (e.g. Bezier curves, object part decomposition) with minimal changes. This research opens up new directions in representing 3D scenes as language, bringing the 3D reconstruction community closer to the recent advances in large language models such as GPT-class models \cite{openai2023gpt4}.

%
%
\bibliographystyle{splncs04}
\bibliography{main}

\clearpage
\appendix

\section{\DatasetName{}}
\label{sec:input_data}

\begin{figure*}[ht]
    \centering
    \includegraphics[width=0.24\linewidth]{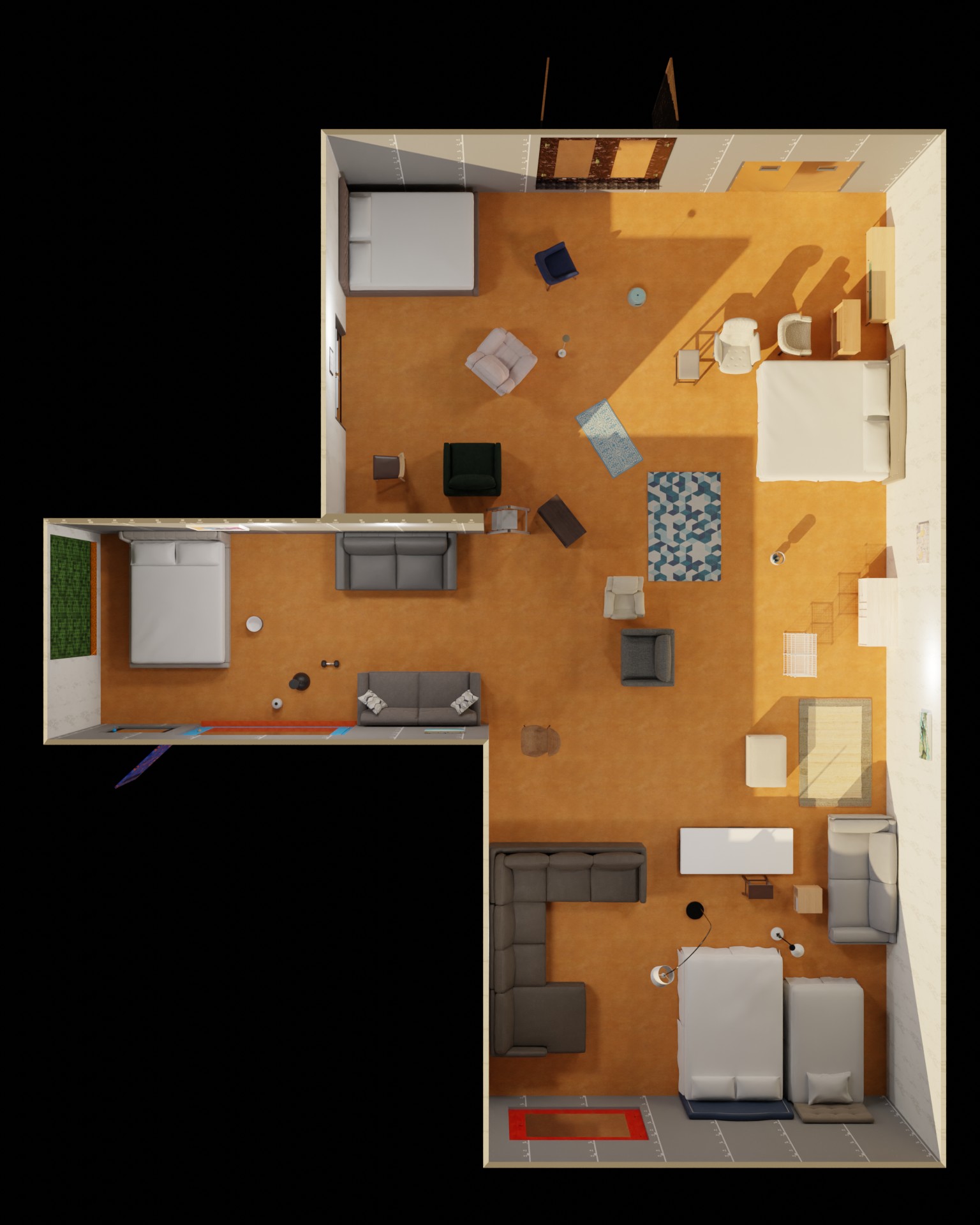} 
    \includegraphics[width=0.24\linewidth]{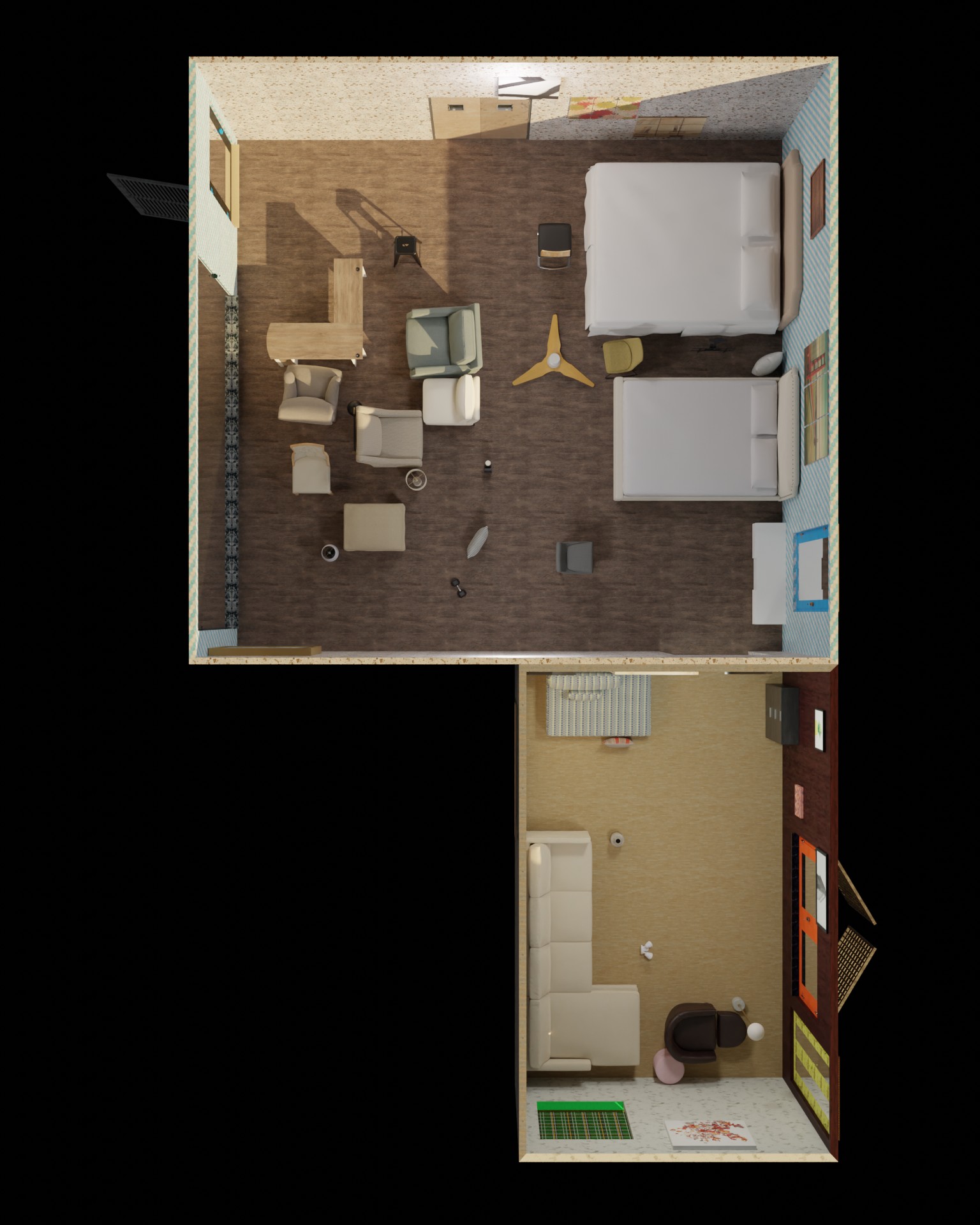} 
    \includegraphics[width=0.24\linewidth]{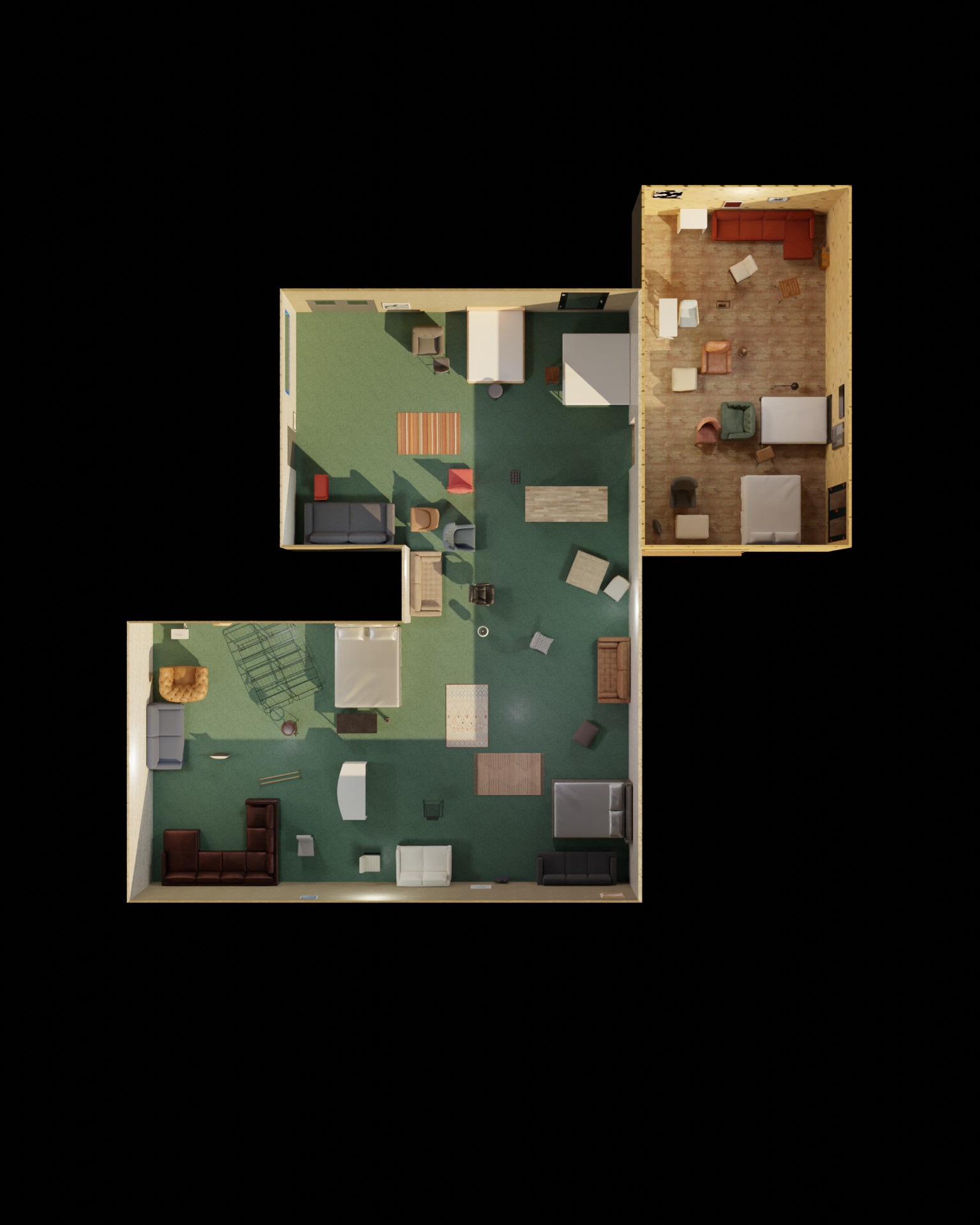} 
    \includegraphics[width=0.24\linewidth]{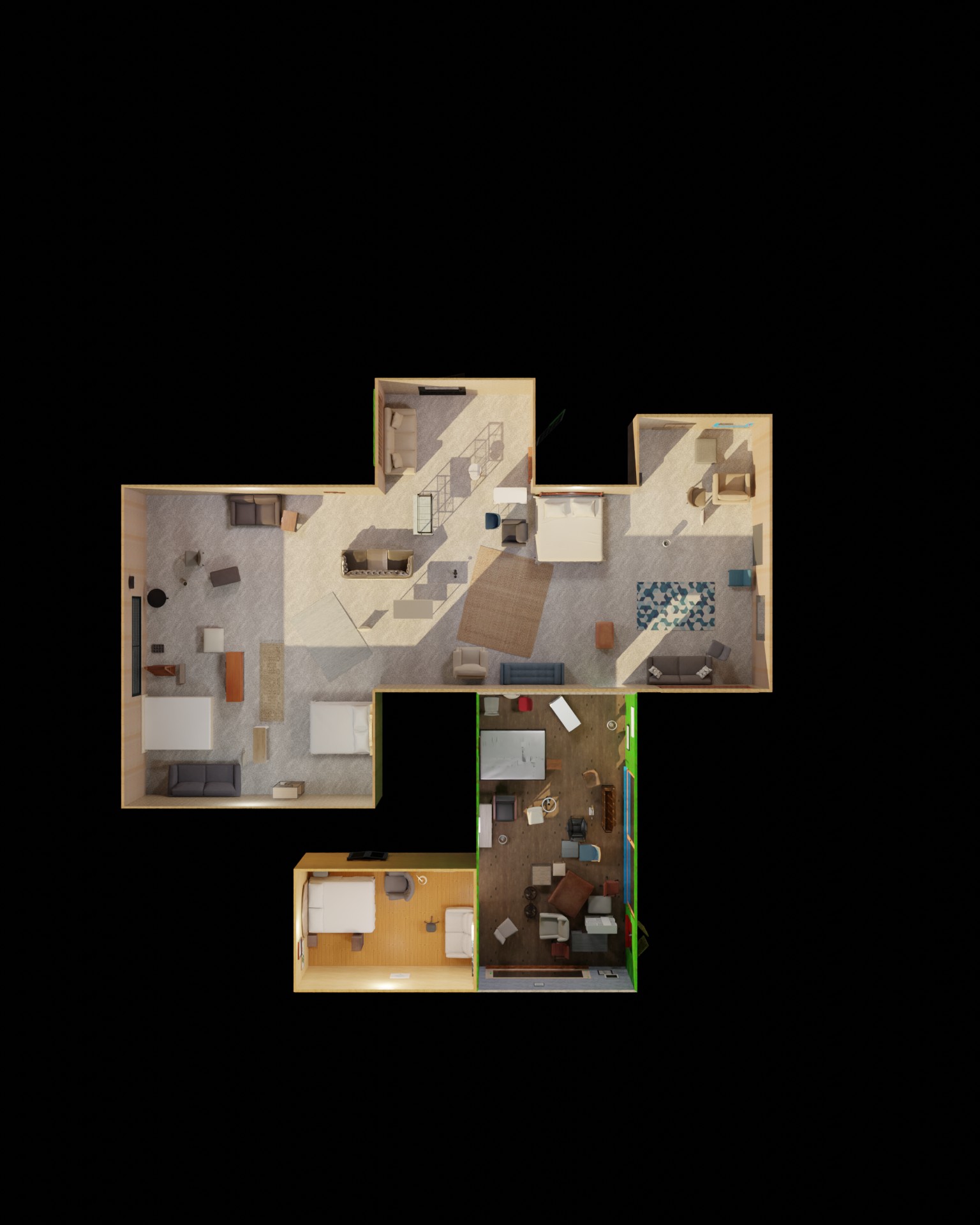}
    \includegraphics[width=0.24\linewidth]{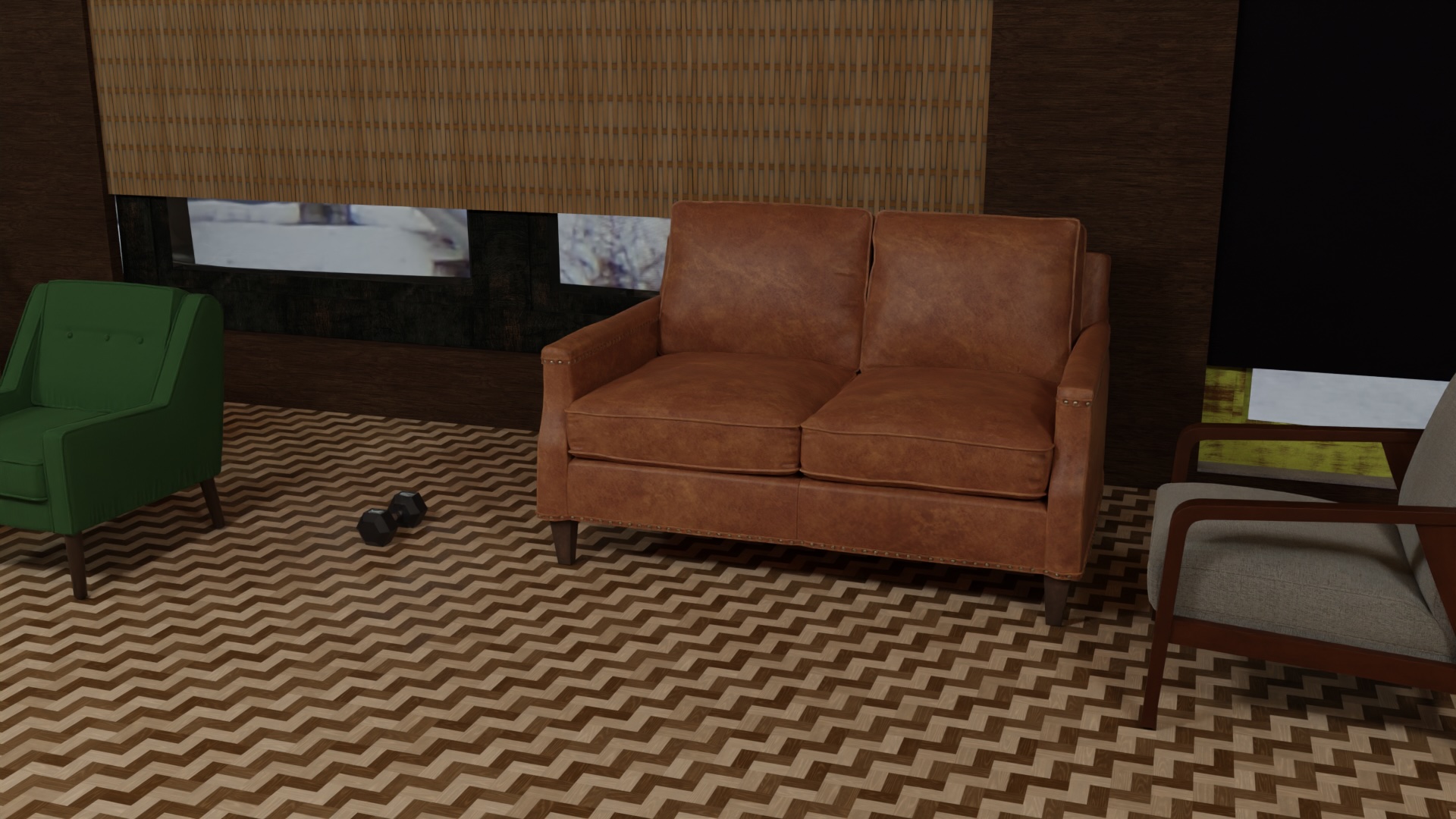} 
    \includegraphics[width=0.24\linewidth]{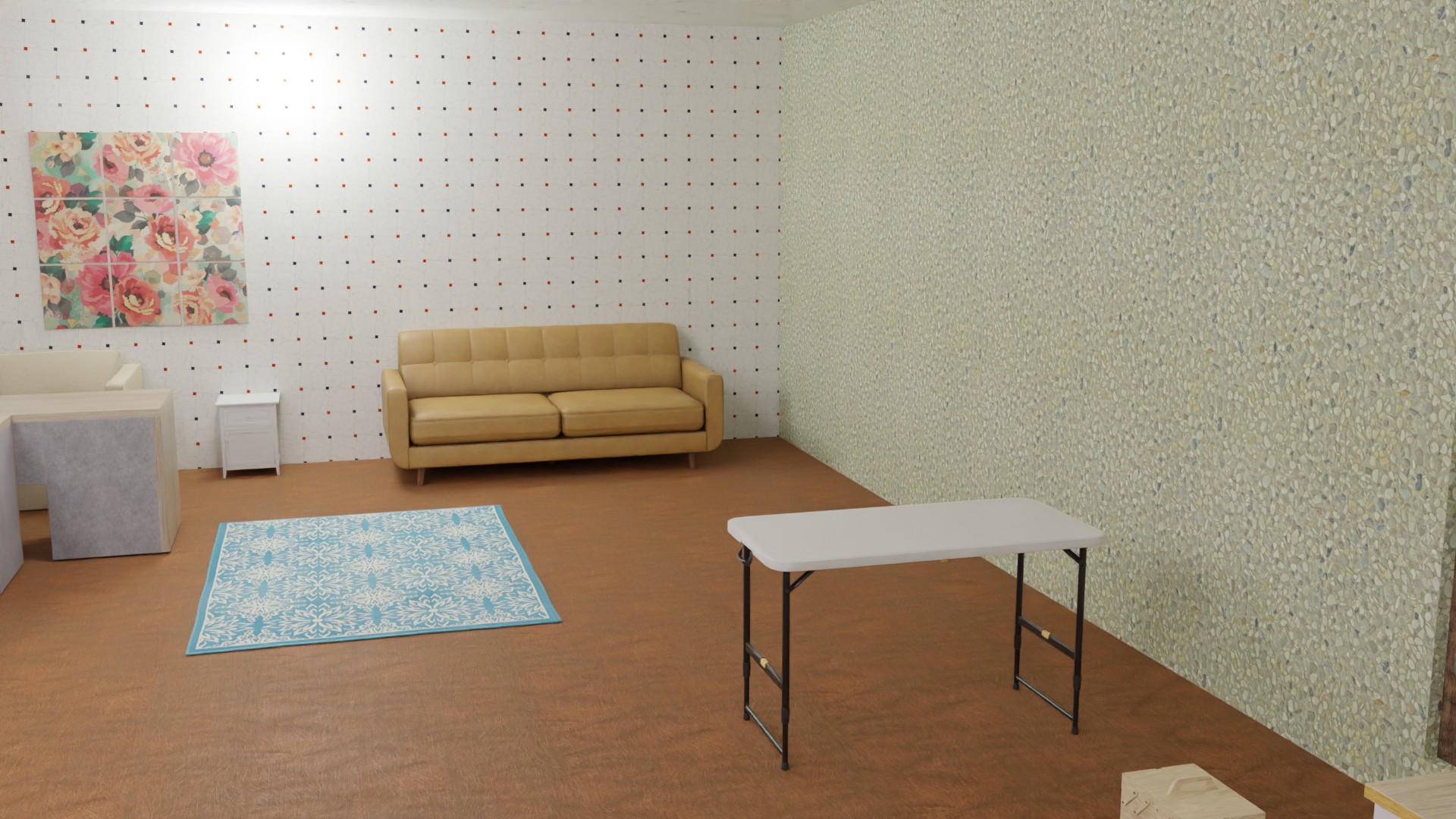} 
    \includegraphics[width=0.24\linewidth]{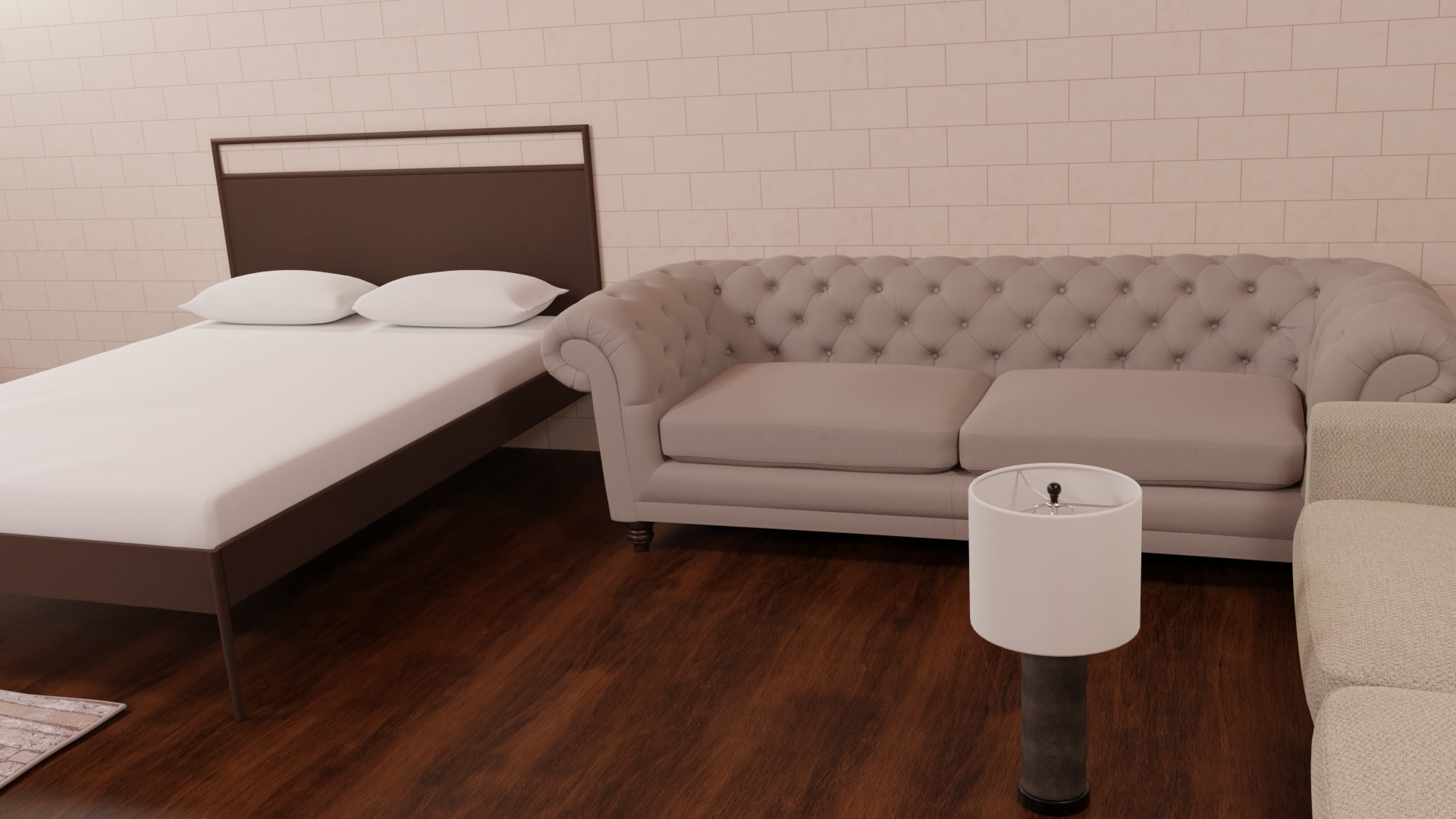} 
    \includegraphics[width=0.24\linewidth]{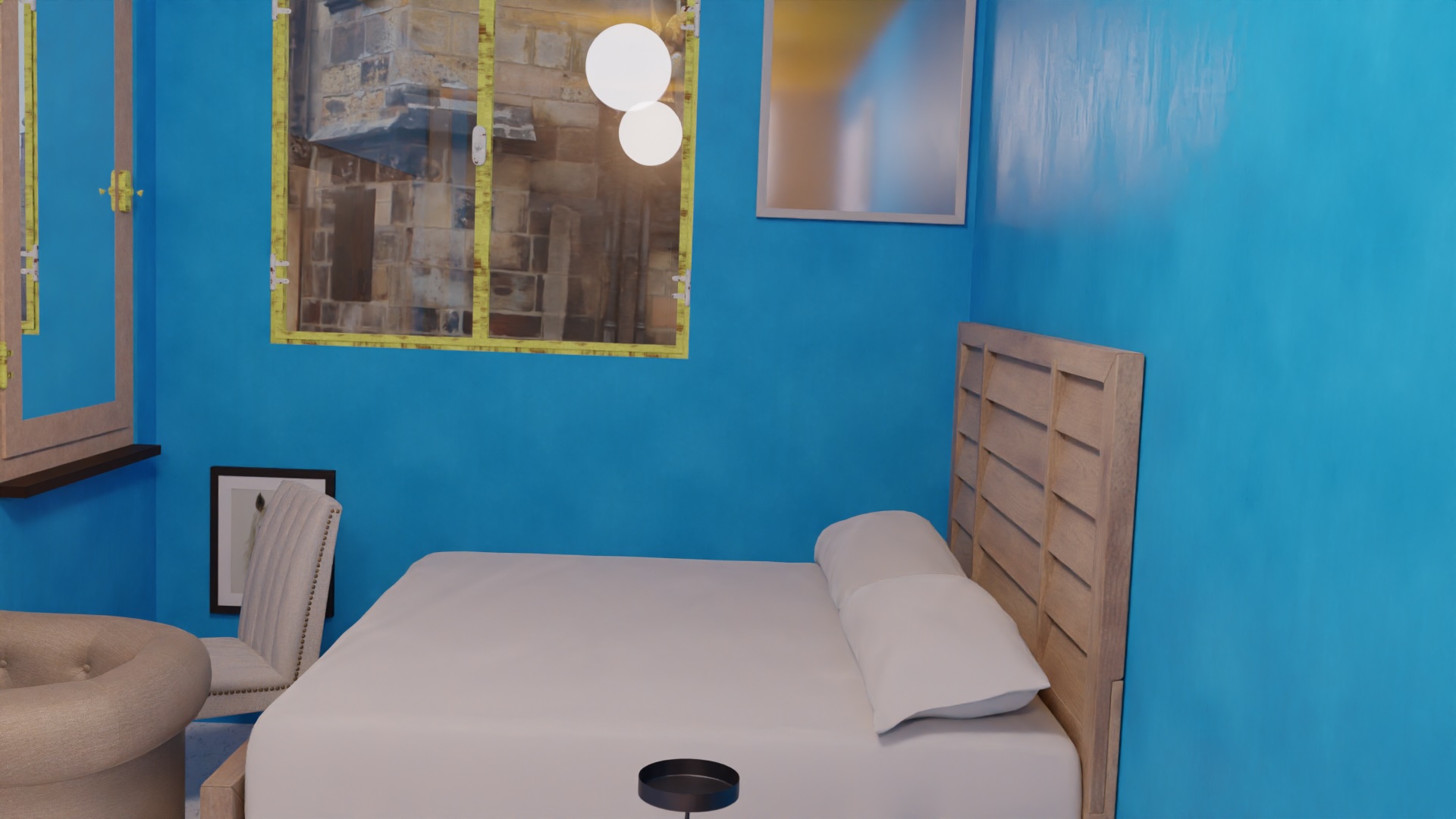}
    \caption{Randomly selected scenes from \DatasetName{}.
        (top) Birds eye view renderings 
        demonstrating room layouts and furniture arrangements.
        (bottom) Ego-centric close-up renderings showing scene details.
    }
    \label{fig:dataset_examples}
\end{figure*}
%

\subsection{Large Scale Training Dataset}

\DatasetName{} consists of $100k$ training pairs
with photo-realistically rendered indoor scenes
coupled with structured language commands.
In addition to these training sequences, \DatasetName{} also provides an additional $1k$ scenes for testing.
Figure~\ref{fig:dataset_examples} presents example scenes from the dataset.
To the best of our knowledge, this is the largest synthetically generated and annotated dataset to date.
%

\begin{figure}
\includegraphics[width=1.0\columnwidth]{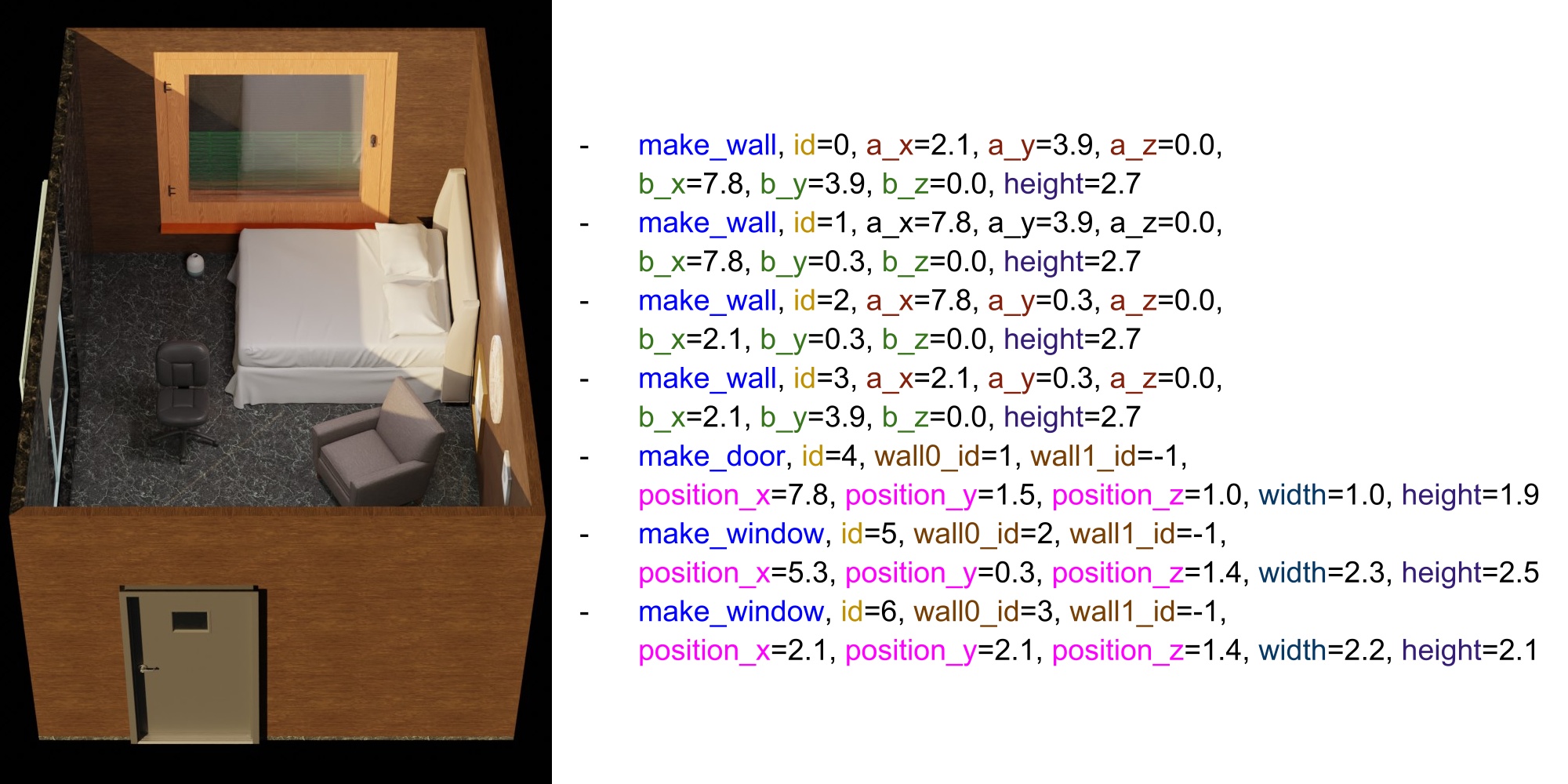}
\caption{Example of training data pair. A scene with objects is shown on the left, while the respective GT \METHOD~language description is shown on the right.}
\label{fig:training_datum}
\end{figure}

Specifically, a training pair for \METHOD~consists of
a 3D scene model represented through a rendered video sequence
(input)
and associated with a sequence of commands (ground truth).
An example training pair for our method is shown in Figure~\ref{fig:training_datum}.

\paragraph{Generation.} Each of our synthetic indoor scenes are generated in the following way:
\begin{enumerate}
    \item Start with:
        \begin{itemize}
            \item A floor plan defining the layout of all rooms
            \item A corresponding 3D model (room \& object geometry, materials and illumination)
        \end{itemize}
    \item A trajectory is estimated going through the 3D scene model, simulating an agent walking around wearing an Aria sensor
    \item A photo-realistic RGB rendering of the trajectory is generated with additional annotation data (depth and instance segmentation maps)
    \item A \SLAM-based point cloud inferred from the synthetic video sequence and aligned with the initial 3D scene model/floor plan
\end{enumerate}


\begin{table}
\small
\begin{centering}
\caption{
    Comparison between existing indoor datasets.
    P-R: Photo-realistic;
    Ego: Ego-centric;
    Camera: Either Fisheye(F) or Pinhole(P), or panorama photo(360);
    Seg: object and layout segmentations rendered to images;
    L-GT: Layout entity ground-truth,
    such as individual wall/window/door parameters.
}
\label{tab:ase_stats}
\resizebox{\linewidth}{!}{%
\begin{tabular}{|c|c|c|ccccccc|c|}
\hline 
Type & Name & Scenes & Trajectory & P-R & Ego & Camera & Depth & Seg & L-GT & License\tabularnewline
\hline 
\hline 
\multirow{7}{*}{Syn} & ASE (\textbf{Ours}) & 100k & \textcolor{green}{$\checkmark$} & \textcolor{green}{$\checkmark$} & \textcolor{green}{$\checkmark$} & F & \textcolor{green}{$\checkmark$} & \textcolor{green}{$\checkmark$} & \textcolor{green}{$\checkmark$} & Agreement needed\tabularnewline
\cline{2-11} \cline{3-11} \cline{4-11} \cline{5-11} \cline{6-11} \cline{7-11} \cline{8-11} \cline{9-11} \cline{10-11} \cline{11-11} 
 & ProcTHOR & 10k & \textcolor{red}{$\times$} & \textcolor{red}{$\times$} & \textcolor{green}{$\checkmark$} & P & \textcolor{green}{$\checkmark$} & \textcolor{green}{$\checkmark$} & \textcolor{green}{$\checkmark$} & Apache2.0\tabularnewline
\cline{2-11} \cline{3-11} \cline{4-11} \cline{5-11} \cline{6-11} \cline{7-11} \cline{8-11} \cline{9-11} \cline{10-11} \cline{11-11} 
 & HyperSim & 461 & \textcolor{green}{$\checkmark$} & \textcolor{green}{$\checkmark$} & \textcolor{green}{$\checkmark$} & P & \textcolor{green}{$\checkmark$} & \textcolor{green}{$\checkmark$} & \textcolor{red}{$\times$} & Special license\tabularnewline
\cline{2-11} \cline{3-11} \cline{4-11} \cline{5-11} \cline{6-11} \cline{7-11} \cline{8-11} \cline{9-11} \cline{10-11} \cline{11-11} 
\cline{2-11} \cline{3-11} \cline{4-11} \cline{5-11} \cline{6-11} \cline{7-11} \cline{8-11} \cline{9-11} \cline{10-11} \cline{11-11} 
 & Structured3D & 3,500 & \textcolor{red}{$\times$} & \textcolor{green}{$\checkmark$} & \textcolor{green}{$\checkmark$} & P & \textcolor{green}{$\checkmark$} & \textcolor{green}{$\checkmark$} & \textcolor{green}{$\checkmark$} & MIT\tabularnewline
\cline{2-11} \cline{3-11} \cline{4-11} \cline{5-11} \cline{6-11} \cline{7-11} \cline{8-11} \cline{9-11} \cline{10-11} \cline{11-11} 
 & SceneNet RGB-D & 57 & \textcolor{green}{$\checkmark$} & \textcolor{green}{$\checkmark$} & \textcolor{green}{$\checkmark$} & P & \textcolor{green}{$\checkmark$} & \textcolor{green}{$\checkmark$} & \textcolor{red}{$\times$} & Special license\tabularnewline
\cline{2-11} \cline{3-11} \cline{4-11} \cline{5-11} \cline{6-11} \cline{7-11} \cline{8-11} \cline{9-11} \cline{10-11} \cline{11-11} 
 & InteriorNet & 10k/1.7M & \textcolor{green}{$\checkmark$} & \textcolor{green}{$\checkmark$} & \textcolor{green}{$\checkmark$} & F, P & \textcolor{green}{$\checkmark$} & \textcolor{green}{$\checkmark$} & \textcolor{red}{$\times$} & Agreement needed\tabularnewline
\hline 
\multirow{3}{*}{Real} & Zillow Indoor & 2,564 & \textcolor{red}{$\times$} & \textcolor{green}{$\checkmark$} & \textcolor{red}{$\times$} & 360 & \textcolor{red}{$\times$} & \textcolor{red}{$\times$} & \textcolor{green}{$\checkmark$} & Apache 2.0\tabularnewline
\cline{2-11} \cline{3-11} \cline{4-11} \cline{5-11} \cline{6-11} \cline{7-11} \cline{8-11} \cline{9-11} \cline{10-11} \cline{11-11} 
 & HM3D & 1,000 & \textcolor{red}{$\times$} & \textcolor{green}{$\checkmark$} & \textcolor{red}{$\times$} & P & \textcolor{green}{$\checkmark$} & \textcolor{red}{$\times$} & \textcolor{red}{$\times$} & MIT\tabularnewline
\cline{2-11} \cline{3-11} \cline{4-11} \cline{5-11} \cline{6-11} \cline{7-11} \cline{8-11} \cline{9-11} \cline{10-11} \cline{11-11} 
 & ScanNet & 1,513 & \textcolor{green}{$\checkmark$} & \textcolor{green}{$\checkmark$} & \textcolor{red}{$\times$} & P & \textcolor{green}{$\checkmark$} & \textcolor{green}{$\checkmark$} & \textcolor{red}{$\times$} & Special license (data)\tabularnewline
\hline 
\end{tabular}
}
\par\end{centering}
\end{table}

\paragraph{Dataset Properties.} An overview of properties of existing indoor datasets is given in Table \ref{tab:ase_stats}.
Note that
we define ego-centric as wearing a real camera on the head or chest
or having a synthetic camera with similar trajectory.
In particular, 
we follow the Aria ego-centric specifications \cite{aria_white_paper}.
Also note that
\emph{InteriorNet} contains 15k sequences 
rendered from 10k randomly selected layouts and
5M images from 1.7M randomly selected layouts,
with 3 images per layout.
\DatasetName{} is especially useful for machine learning tasks not just due to its sheer scale (allowing algorithms to generalize well), but also the wide variety of available ground truth annotations.
For example, the layout ground truth (L-GT) is critical for training \METHOD,
but not included in the majority of other datasets.

\DatasetName{} is made publicly available to the research community.
Users must agree to a standard licence to prevent data misuse
and it is to be used for non-commercial purposes only.

\subsection{Semi-dense Point Clouds from Video Sequences}

We utilize the open-source Machine Perception Services (MPS) from Project Aria~\cite{AriaMPS2023} to estimate a SLAM trajectory and generate a point cloud map of the scene. Similarly to LSD-SLAM \cite{engel14eccv} they maximize the extracted geometric information from a video sequence by estimating points for all image regions with non-negligible gradient. Each point is parameterized by an inverse distance and its associated Gaussian uncertainty in the frame in which it is first observed. KLT-based \cite{tomasi1991detection} epipolar line-searches in subsequent frames provide sub-pixel accurate short and large-baseline measurements that are absorbed using a Kalman filter update. While points are associated with a final estimated uncertainty, they consider utilizing this information in a probabilistically-sound way as beyond the scope of their work, and instead choose to sub-select points whose uncertainty is below a predefined threshold.

%

\section{Structured Language Commands}
\label{sec:parameters_supp}

Command parameters can have data types such as \lstinline[style=dtypestyle]!float! or \lstinline[style=dtypestyle]!int!. The full list of parameters for each command can be found in Table~1 of the main paper. Below, we provide detailed descriptions of each parameter:

\begin{itemize}
\item Wall parameters
    \begin{itemize}
    \small
        \item \lstinline[style=cmdstyle]!id!: The ID of the wall.
        \item \lstinline[style=cmdstyle]!(a_x,a_y,a_z) / (b_x,b_y,b_z)!: The $x,y,z$ coordinates of the first / second corner of the wall.
        \item \lstinline[style=cmdstyle]!height!: The height of the wall. Note that we assume the walls are straight and gravity-aligned.
    \end{itemize}
    
\item Door/Window parameters
    \begin{itemize}
    \small
        \item \lstinline[style=cmdstyle]!id!: The ID of the door/window.
        \item \lstinline[style=cmdstyle]!wall0_id,wall1_id!: The IDs of the (potentially two) walls that a door/window is attached to.
        \item \lstinline[style=cmdstyle]!position_x,position_y,position_z!: The $x,y,z$ coordinates of the centre of the door/window.
        \item \lstinline[style=cmdstyle]!width, height!: The width and height of the door/window.
    \end{itemize}
\end{itemize}

\section{Network Architectures}
\label{sec:architecture}
\subsection{Point Cloud Encoder}
\label{subsec:point_cloud_encoder}

The point cloud encoder is essentially a ResNet-style~\cite{he2016deep} encoder that employs sparse 3D convolutions~\cite{tang2020searching,tang2022torchsparse} in place of standard 3D convolutions. It uses a total of 5 down convolutional layers with a kernel size of 3 and a stride of 2. This architecture effectively reduces the number of points (i.e. active sites) by $\approx 1000$. As a result, the feature sizes are manageable in terms of size and can be used effectively in the subsequent Transformer decoder~\cite{vaswani2017attention}. The point cloud encoder consists of $\approx 20M$ optimizable parameters, which contribute to its capacity and ability to capture intricate geometric information.

\subsection{Transformer Decoder}
\label{app:decoder}

Our implementation includes a transformer decoder consisting of 8 layers, each with 8 heads for multi-head attention, and a feature dimension of $512$. This configuration results in a relatively modest set of $\approx 35M$ parameters. Our vocabulary size is $2048$, which we also use as the maximum sequence length of tokens. While we could theoretically increase the vocabulary size to accommodate a larger number of tokens, in practice the majority of the released rendered scenes can be accurately represented using significantly fewer tokens than $2048$. 

In some scenarios, we employ nucleus sampling~\cite{holtzman2019curious} with a top-p probability mass during autoregressive inference. By using nucleus sampling, we limit the selection of next tokens to a subset with a cumulative probability threshold, allowing for greater exploration at inference time. Quantitative results are decoded greedily.

\section{Training Methodology}

We use the AdamW optimizer 
with a default initial learning rate of $10^{-4}$ and
weight decay as well as dropout enabled.
For the image-only Raytran-based encoder model~\cite{tyszkiewicz2022raytran},
we found that an initial learning rate of $10^{-3}$ provided better convergence.
We train all our methods with an effective batch size of 64, which may be distributed across multiple nodes. 
During training, the only augmentations we perform are: 1) up to 360 degrees of rotation around the z-axis (since the scenes are assumed to be gravity-aligned), and 2) random subsampling of point clouds up to a specificed maximum number of points (500k). 
Training times lasted approximately between 3 and 4 days. 

Our training loss is the standard cross-entropy loss on next token prediction, similar to how most LLMs are trained ~\cite{openai2023gpt4,sutskever2014sequence}.

\section{Tokenization Details}

The conversion of a parameter $\textbf{\texttt{x}}$ into an integer token $\textbf{\texttt{t}}$ is determined by two factors: its data type and its magnitude. In general, parameters with \lstinline[style=dtypestyle]!int! and \lstinline[style=dtypestyle]!bool! data types are converted using the $\textbf{\texttt{t = int(x)}}$ operation. For \lstinline[style=dtypestyle]!float! parameters, the conversion involves $\textbf{\texttt{t = round(x/res)}}$, where $\textbf{\texttt{res}}$ represents the resolution. 
Note that  by designing the \METHOD~language,
we also design the tokenization.
This is notably different from standard NLP tokenization,
which involves \emph{B}yte-\emph{P}air \emph{E}ncodings (\emph{BPE}) \cite{openai2023gpt4}.

\section{Additional Results: Layout Prediction}
In this section,
we present additional qualitative results,
comparisons with related works and
more evaluations with respect to extending \METHOD{}.

\begin{figure}[t]
    \centering
    \includegraphics[width=\columnwidth]{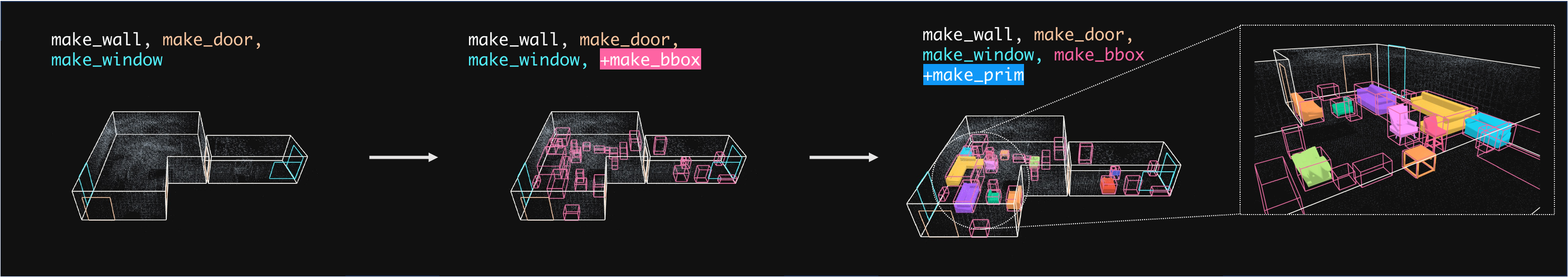}
    \caption{
        An illustrative example of how the expressiveness of SceneScript's reconstruction increases through the addition of new commands.
        (left) Layout commands only: walls, doors and windows.
        (middle, left) Addition of \texttt{make\_bbox} enriches the scene reconstruction with bounding boxes for detected objects.
        (middle, right) Addition of \texttt{make\_prim} adds volumetric primitives for detected chairs, sofas and tables.
        (right) Close-up illustrating the fidelity possible with just these five commands.
    }
    \label{fig:commandVisExamples0}
\end{figure}

\subsection{Visualization of High-Level Commands}

Figure \ref{fig:commandVisExamples0} presents a visual example of how the fidelity of \METHOD's reconstruction can be increased through the addition of new commands. 
Initially, structural room layouts are represented by three commands: \texttt{make\_wall}; \texttt{make\_door}; and \texttt{make\_window}.
Just through the addition of \texttt{make\_bbox}, scene content is now present in the reconstruction in the form of object detections.
Finally for the commands discussed in the main paper, \texttt{make\_prim} for the three selected target classes enables not just the capture of the scene's overall structure and content, but also much finer reconstruction of scene objects themselves. 

Importantly, each of these levels of detail is enabled without change to \METHOD's network architecture, and instead just by increasing the expressiveness of the structured language it infers.

Note that
the volumetric primitive commands for detected objects are a proof of concept.
We trained our models for the object primitive commands only on a subset
of the available object types from the \DatasetName{}.
Supported object class labels are ``\emph{chair}'', ``\emph{sofa}'' and ``\emph{table}''.
Objects with these labels are modeled by cuboid and cylinder primitives.
Detected bounding boxes of object instances with unsupported classes remain empty.

\begin{figure}[t]
    \centering
    \includegraphics[width=0.5\columnwidth]{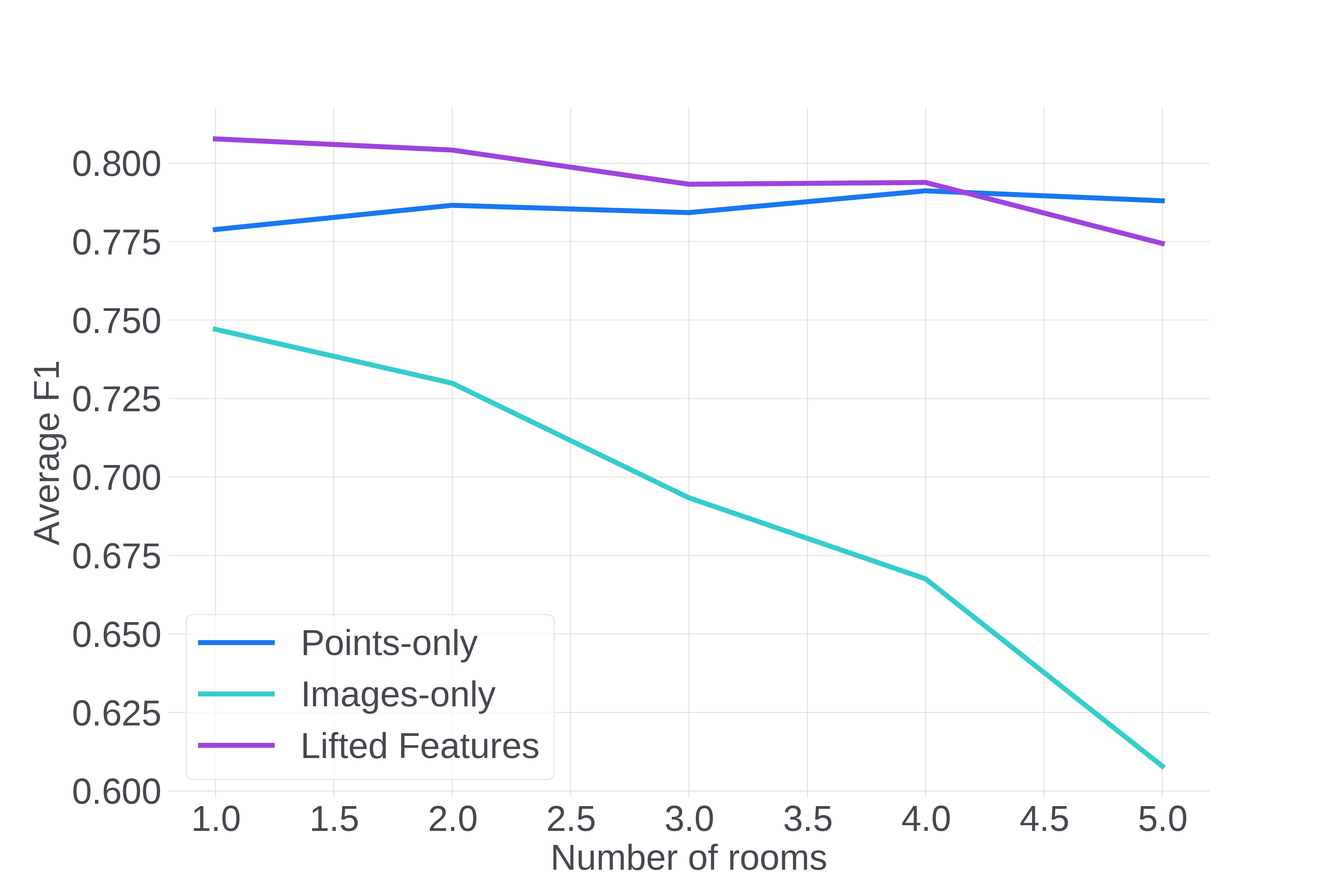}
    \caption{
        F1-Score model performance graphs
        for our various encoder variants
        as functions of the number of rooms in a scene.
    }
    \label{fig:roomsf1}
\end{figure}
\label{subsec:failure_cases}
\begin{figure}[h]
    \centering
    \includegraphics[width=0.8\columnwidth]{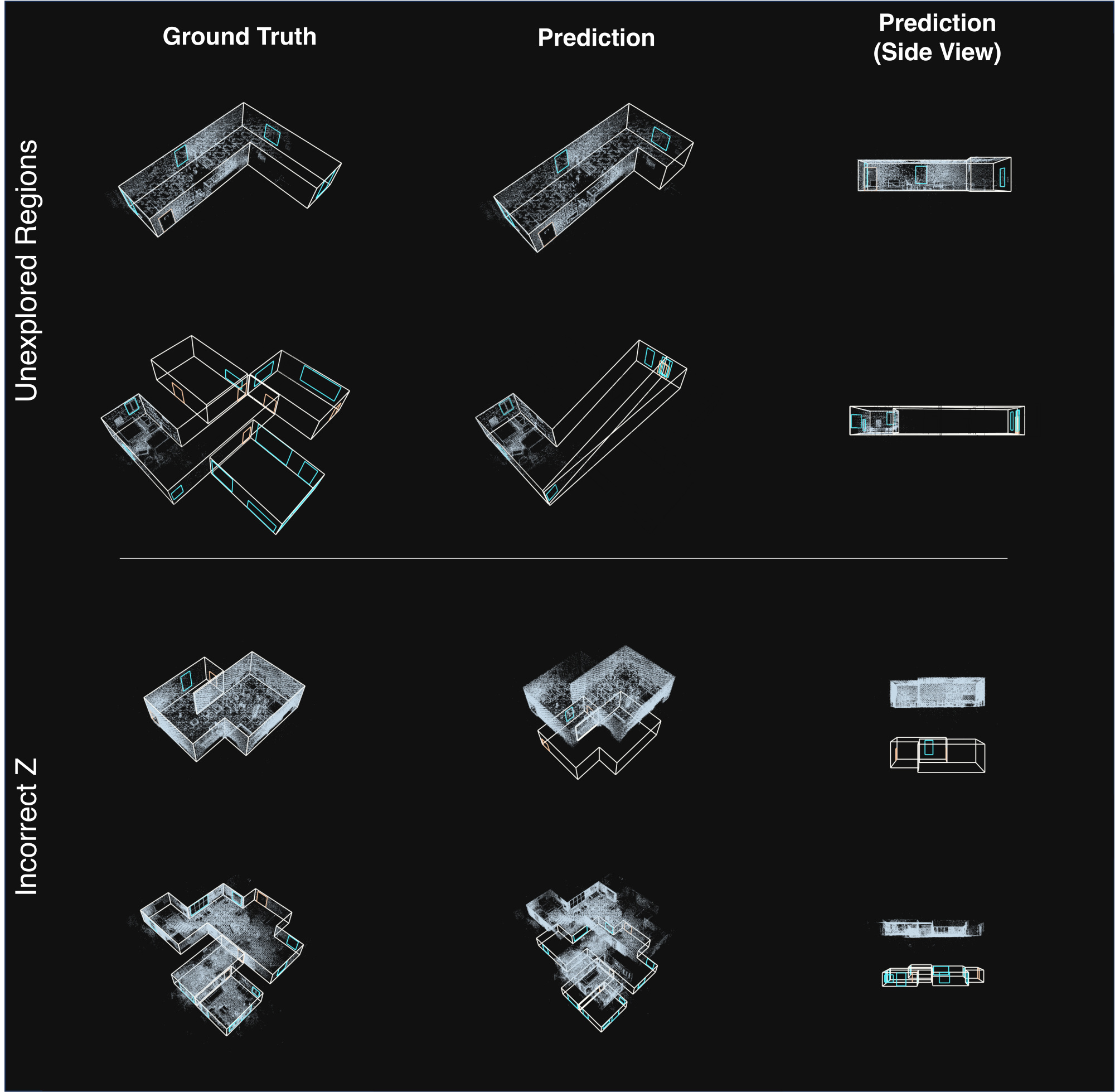}
    \caption{Examples from two notable failure types observed in \METHOD predictions. (top) Limited exploration of the scene makes the ground truth difficult, or in some cases potentially impossible to predict. (bottom) Accurate overall room structure is predicted, but at an incorrect Z value.}
    \label{fig:failures}
\end{figure}

\subsection{Model Performance with respect to Scene Complexity}
\label{app:num_rooms}

The Average F1-score graphs of Figure \ref{fig:roomsf1} demonstrate
the performance of our \METHOD{} model with varying encoders
as a function of the number of rooms in a scene.
Our \METHOD{} model performs constantly well 
when inputting points only or lifted features.
As opposed to this, 
performance drops drastically with increasing room number
when encoding scenes only using images.
We posit that the decrease in performance is due to
the model's lack of occlusion reasoning.
With increasing number of layout elements,
the rays linked to image observations traverse
more scene space by going
through a higher number of rooms
when ignoring wall intersections.
This likely results in 
our model falsely attending to occluded image observations.

\subsection{Failure Cases}

In this section, we detail observed failure types for the task of layout estimation on \DatasetName{}.
Aside from expected errors such as slightly incorrect wall corner, window and door placement, or entirely missed, we observe two notable failure modes for \METHOD.

The more common of the two occurs due to non-complete exploration of the target scene. In this scenario there are significant portions of the scene structure that are poorly observed, potentially not at all, making the ground truth structure near unpredictable.

An especially interesting failure mode is the reconstruction of accurate room structure, but at an incorrect Z-value. For the point cloud-based encoder configurations, we suspect that this failure mode is caused by particular sensitivity to noise to point outliers in the Z-direction. This failure mode is also observed in the image-only encoder configuration, suggesting it also exhibits more sensitivity to in the Z direction than XY.

We visualize a couple of examples for each of these failure types in Figure~\ref{fig:failures}. Worth noting is that this figure is comprized of scenes taken from the worst 10 predictions out of the 1000 scene test set, as defined by wall entity distance. Therefore, while clearly illustrating the failures described, they should not be taken as representative of general prediction quality.

\begin{table}
\centering
    \caption{
    Accuracy reported as raw entity distance for the encoder setups tested for \METHOD. 
    }
\begin{tabular}{c|cc|cc|cc}
        & \multicolumn{6}{c}{Entity Distance (cm)} \\
        \hline
         & \multicolumn{2}{c|}{Wall} & \multicolumn{2}{c|}{Door} & \multicolumn{2}{c}{Window}\\
        Method & med. & p90 & med. & p90 & med. & p90 \\
        \hline \hline
        Point Cloud &       4.7 & 7.2   & 5.0 & 6.7     & 6.9 & 37.6 \\
        Lifted Features &   4.8 & 7.1   & 4.8 & 6.1     & 5.9 & 26.2  \\
        Image-only &        6.7 & 17.3  & 5.8 & 8.9     & 9.0 & 45.7 \\
    \end{tabular}
    \label{table:plane_distances}
\end{table}
\subsection{Quantitative Evaluation of Layout Predictions}
\label{app:per_entity}

We include an additional breakdown of the entity distance accuracy metrics in Table~\ref{table:plane_distances}. 
This breakdown of accuracy makes apparent that the improvement offered by lifting image features onto the semi-dense point cloud comes largely in the prediction of windows and doors.
Following the same trend as the results included in the main paper, we observe that windows appear to be the most challenging class to predict accurately. 
However in spite of this challenge, the 90th percentile of window predictions falls within 0.5m of the ground truth for all encoder setups tested.

\section{Additional Results: Object Detection}

\subsection{Implementation Details}
\subsubsection{Training Details of \METHOD{}.}
As outlined in the paper, a significant advantage of \METHOD{} lies in its seamless adaptability to other tasks through the addition of new language commands. Here, for instance, we integrate \texttt{make\_bbox} to denote 3D oriented bounding boxes.

Notably, no architectural changes to \METHOD{} have been implemented to facilitate training for object detection. We utilize the point cloud encoder and language decoder detailed in Section \ref{sec:architecture}. The entire training objective is a single cross-entropy loss, which stands as the de facto-standard in training LLMs. The model is trained for $\approx 200k$ iterations using an effective batch size of 64. For this experiment, we only trained a point cloud version of SceneScript.

\subsubsection{Baseline Implementation Details.}

We list implementation details for each method below:

\paragraph{3DETR~\cite{misra2021end}:} We downloaded the weights for both 3DETR and 3DETR-m, trained for 1080 epochs on ScanNet, from the official Github repository. We evaluated both models on each ScanNet validation examples, subsampled to $40k$ points. Predictions were thresholded at probability 0.5. We attemped to run NMS in 3D, but achieved worse results, thus the numbers reported in the main paper do not include NMS.

We trained 3DETR (not 3DETR-m) on \DatasetName{} using almost the same configuration as trained on ScanNet. The differences include: a batch size of 128, a PointNet set aggregation downsampling to 4096 (compared to 2048 for ScanNet), 512 furthest point samples as detection queries (compared to 256 for ScanNet), and 200k points per example. 
    
\paragraph{Cube R-CNN~\cite{brazil2023omni3d}:} This method predicts 3D bounding boxes from a single RGB image. To obtain 3D bounding box predictions for an entire scene, we accumulate per-frame predictions via a matching-based tracker. At a high-level, we match predictions of the same class between frames with Hungarian matching with a cost based on IoU and bounding box distances. Then the final bounding box parameters are computed as a running average of the matched and tracked predictions. For evaluation, the accumulated predicted boxes were thresholded at probability 0.5.

\paragraph{ImVoxelNet~\cite{rukhovich2022imvoxelnet}:} This model predicts 3D bounding boxes from a set of RGB images. We trained this method using 10 consecutive frame snippets from \DatasetName{}. During evaluation, we run the model on overlapping 10-frame snippets and apply the same bounding box tracker as described for Cube R-CNN. For evaluation, the accumulated predicted boxes were thresholded at probability 0.1.
    
\paragraph{SoftGroup~\cite{vu2022softgroup}:} Since this is primarily a 3D semantic instance segmentation method, we extract axis-aligned bounding boxes from the predictions by utilizing the predicted instance masks and computing the \textit{minimum} and \textit{maximum} extents of the point set belonging to each instance. The geometric mean of these extents serves as the box center, and the difference between the maximum and minimum extents provides the box scale. Since the bounding boxes are intended to be axis-aligned, the angle is kept at 0. By combining this information with the predicted semantic class, one can conduct evaluations over 3D bounding boxes. We used a publically available checkpoint provided by the authors to conduct inference and extract bounding boxes for evaluation following the aforementioned procedure. 

Note that for ScanNet~\cite{dai2017scannet}, we use the axis-aligned bounding boxes for ground truth as extracted in~\cite{misra2021end,qi2019deep}.

\subsection{Sparse Encoder with 3DETR Head}
\label{subsec:sparsecnn_threedetr}

\begin{table}[h]
\centering
    \caption{Replacing the 3DETR encoder with a SparseCNN results in better performance on \DatasetName.}
    \begin{tabular}{c|c|cc}
        & & \multicolumn{2}{c}{F1} \\
         Method & Input & @.25 IoU & @.50 IoU \\
        \hline \hline
        3DETR '21~\cite{misra2021end} & Points & 0.201 & 0.078 \\
        SparseCNN~\cite{tang2022torchsparse} + 3DETR~\cite{misra2021end} & Points & 0.381 & 0.191 \\
    \end{tabular}
    \label{tab:sparsecnn_3detr}
\end{table}

We run an experiment that confirms that 3DETR's standard settings are well-suited to ScanNet~\cite{dai2017scannet} and SUN RGB-D~\cite{song2015sun}, but perform poorly on \DatasetName{}. For this experiment, we use the same sparse point cloud encoder that SceneScript uses (see Section~\ref{subsec:point_cloud_encoder}) while using the 3DETR decoder. Similar to the pure 3DETR model trained on \DatasetName{}, we increased the number of detection queries to 512, and used 200k points per example for training. We denote this model as SparseCNN+3DETR. Due to lack of resources, this model was only partially trained.

\begin{figure}[t]
    \centering
    \includegraphics[width=\columnwidth]{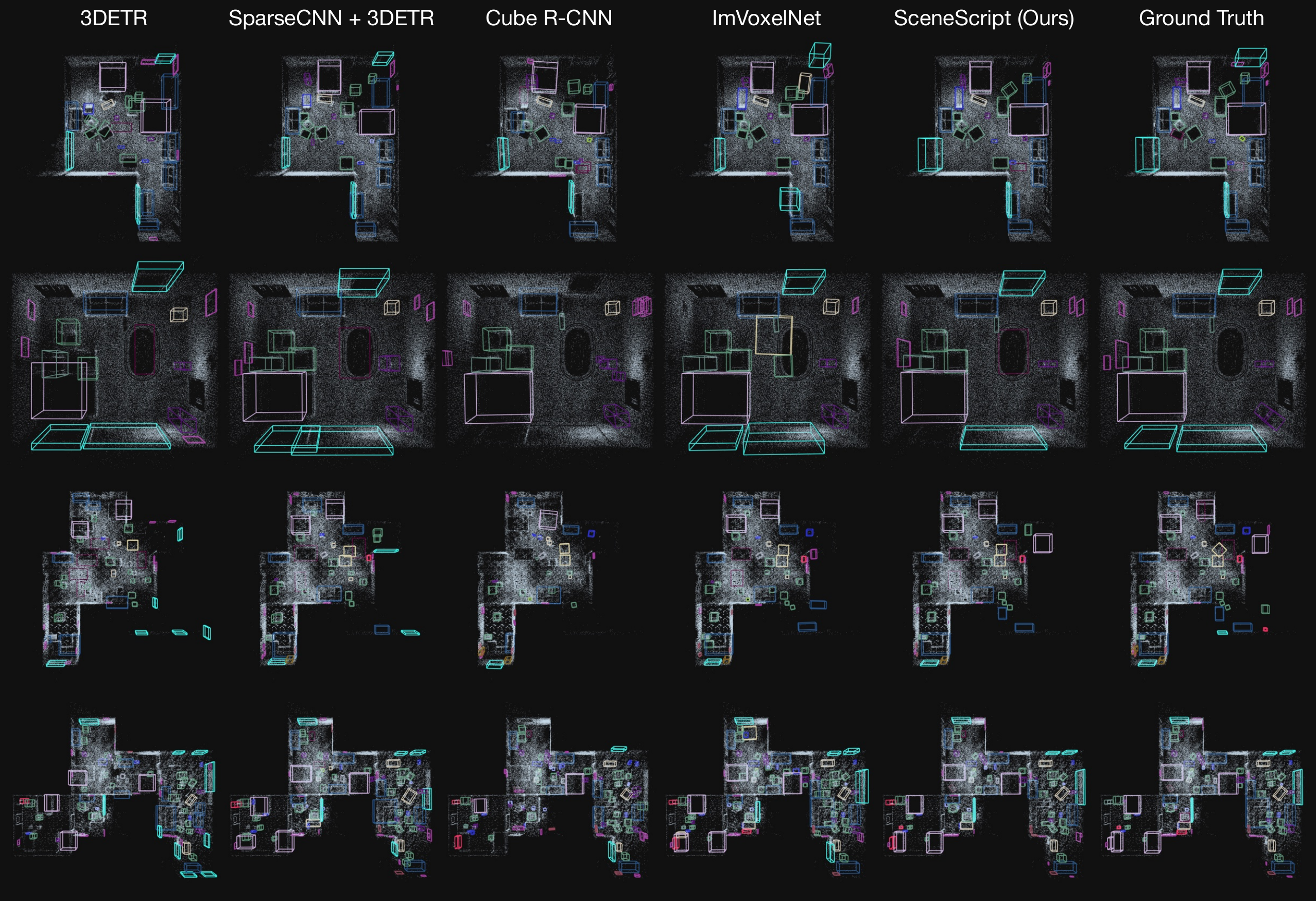}
    \caption{Qualitative results of predicted bounding boxes on \DatasetName{}. Each bounding box is colored by its class. The colors are: \textcolor{table_bbox_color}{table}, \textcolor{sofa_bbox_color}{sofa}, \textcolor{shelf_bbox_color}{shelf}, \textcolor{chair_bbox_color}{chair}, \textcolor{bed_bbox_color}{bed}, \textcolor{floor_mat_bbox_color}{floor\_mat}, \textcolor{exercise_weight_bbox_color}{exercise\_weight}, \textcolor{cutlery_bbox_color}{cutlery}, \textcolor{container_bbox_color}{container}, \textcolor{clock_bbox_color}{clock}, \textcolor{cart_bbox_color}{cart}, \textcolor{vase_bbox_color}{vase}, \textcolor{tent_bbox_color}{tent}, \textcolor{flower_pot_bbox_color}{flower\_pot}, \textcolor{pillow_bbox_color}{pillow}, \textcolor{mount_bbox_color}{mount}, \textcolor{lamp_bbox_color}{lamp}, \textcolor{ladder_bbox_color}{ladder}, \textcolor{fan_bbox_color}{fan}, \textcolor{cabinet_bbox_color}{cabinet}, \textcolor{jar_bbox_color}{jar}, \textcolor{picture_frame_bbox_color}{picture\_frame}, \textcolor{mirror_bbox_color}{mirror}, \textcolor{electronic_device_bbox_color}{electronic\_device}, \textcolor{dresser_bbox_color}{dresser}, \textcolor{clothes_rack_bbox_color}{clothes\_rack}, \textcolor{battery_charger_bbox_color}{battery\_charger}, \textcolor{air_conditioner_bbox_color}{air\_conditioner}, \textcolor{window_bbox_color}{window}.}
    \label{fig:ase_bbox_qualitative}
\end{figure}

\begin{figure}[t]
    \centering
    \includegraphics[width=\columnwidth]{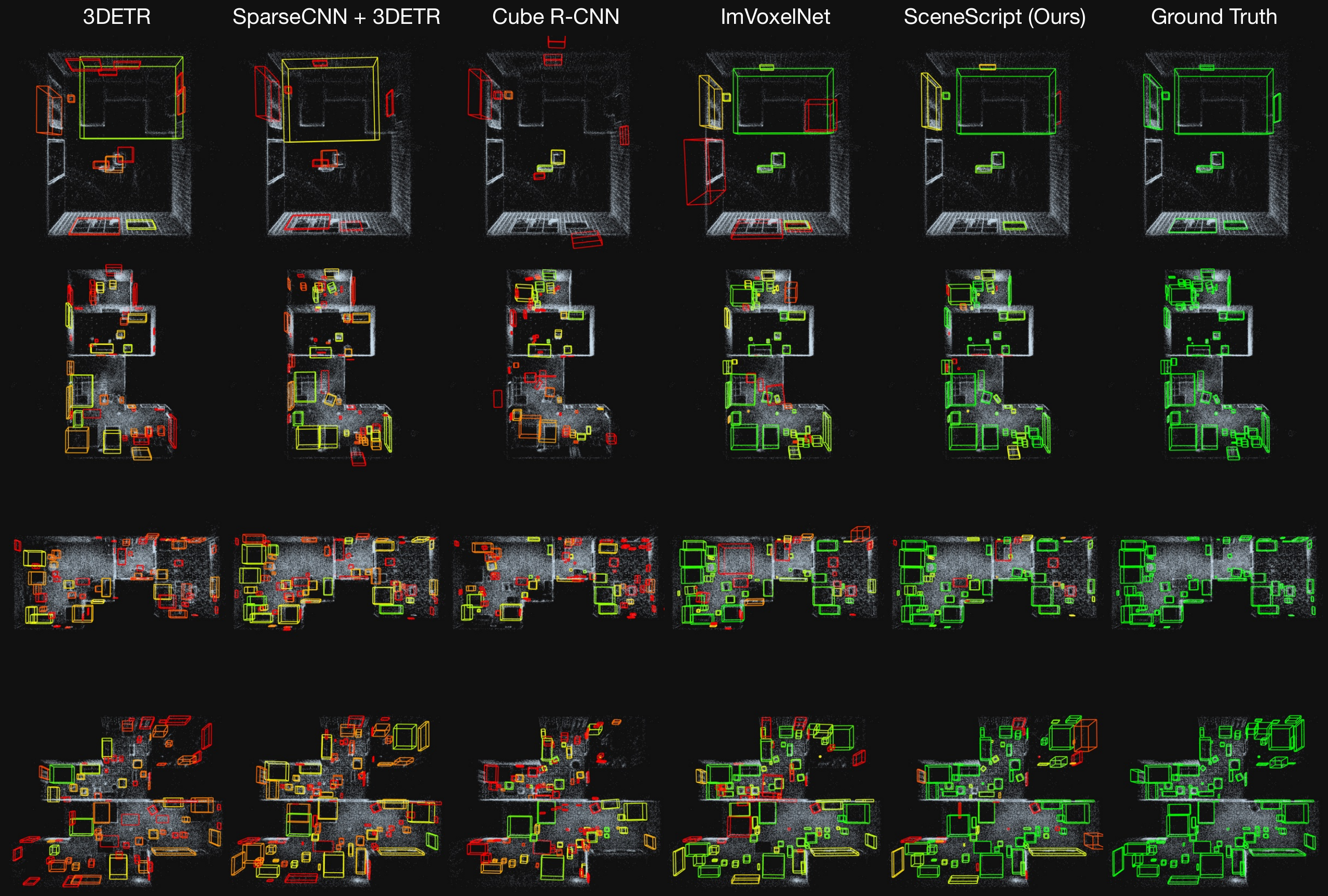}
    \caption{Qualitative results of predicted bounding boxes on \DatasetName{}. Each bounding box is colored by its IoU with its matched ground truth bounding box. The color is interpolated from green (\textcolor{green}{IoU = 1.0}) to yellow (\textcolor{yellow}{IoU = 0.5}) to red (\textcolor{red}{IoU = 0}).}
    \label{fig:ase_bbox_qualitative_iou}
\end{figure}

In Table~\ref{tab:sparsecnn_3detr}, we show that replacing 3DETR's Transformer encoder with a sparse CNN encoder~\cite{tang2022torchsparse,tang2020searching} results in stronger performance. We hypothesis that this is due to the non-uniformity of the point clouds arising from Project Aria's semi-dense point clouds from its Machine Perception Services~\cite{AriaMPS2023}. The first two columns of Figures~\ref{fig:ase_bbox_qualitative} and~\ref{fig:ase_bbox_qualitative_iou} qualitatively demonstrates more accurate predictions with this encoder replacement.

\subsection{Qualitative Results on \DatasetName{}}

In Figure~\ref{fig:ase_bbox_qualitative}, we show qualitative results of all the methods trained on \DatasetName{}. This figure demonstrates the difficult of predicting objects in \DatasetName{} as it is very cluttered. Also, due to the generated trajectories not necessarily visiting every part of the scene, some ground truth bounding boxes have very little points associated with them (see the ground truth in row 3. The lower right corner has very few points yet there are bounding boxes present). 

Most methods tend to correctly predict the larger categories (e.g. \textcolor{bed_bbox_color}{bed} and \textcolor{sofa_bbox_color}{sofa}). However, the small object categories (e.g. \textcolor{jar_bbox_color}{jar} and \textcolor{flower_pot_bbox_color}{flower\_pot}) are much harder to detect, thus the ground truth for these categories typically have 0 IoU with predictions (see Figure~\ref{fig:ase_bbox_qualitative_iou} for qualitative predictions visualised with IoU scores). This leads to relatively low F1 scores for some of the baselines (e.g. 3DETR) due to averaging the F1 scores across classes, while visually the predictions look relatively reasonable. We also include results from SparseCNN+3DETR (details can be found in Section~\ref{subsec:sparsecnn_threedetr}). It can be seen from Figures~\ref{fig:ase_bbox_qualitative} and~\ref{fig:ase_bbox_qualitative_iou} that it qualitatively performs better on \DatasetName{} than a pure 3DETR model.

\begin{table}[h]
\centering
    \caption{mAP for baselines trained on \DatasetName.}
    \begin{tabular}{c|c|cc}
        & & \multicolumn{2}{c}{mAP} \\
         Method & Input & @.25 IoU & @.50 IoU \\
        \hline \hline
        3DETR '21~\cite{misra2021end} & Points & 0.148 & 0.040 \\
        SparseCNN~\cite{tang2022torchsparse} + 3DETR~\cite{misra2021end} & Points & 0.308 & 0.115 \\
        Cube R-CNN '23~\cite{brazil2023omni3d} & RGB & 0.383 & 0.181 \\
        ImVoxelNet '22~\cite{rukhovich2022imvoxelnet} & RGB & 0.648 & 0.572\\
    \end{tabular}
    \label{tab:baseline_map}
\end{table}

\subsection{mAP Metrics for Baselines trained on \DatasetName{}}
\label{app:baseline_map}

In Table~\ref{tab:baseline_map}, we list the mAP values for methods trained on \DatasetName{}.

\subsection{Discussion of Average Precision Metric}
Average precision (AP) has become a standard metric for measuring 3D object detection performance. A general outline of the procedure required to calulate this metric is to collect detections across a number of scenes, rank each by descending confidence. Average precision is then computed from this detection pool by framing it as an information retrieval task: order of retrieval determined by the confidence ranking; success of a retrieval determined by an IoU threshold (typically 0.25 or 0.5 for 3D object detection). This framing enables the generation of a precicision-recall curve for the detector, with the average precision given by an approximation of the area underneath this curve.

A drawback of this information retrieval framing is that it is order variant, and requires that the relative certainty of detections across scenes be determinable. While many prior detection methods regress a logit that can naturally represent this certainty, \eg the classification logit is often used, \METHOD{}'s detections are more binary: either the object is present in the predicted sequence or not. Within a single scene, we may be able to leverage a heuristic such as sequence order to determine relative certainty, \ie most certain detections appear sooner in the prediction order (although we have not investigated whether this actually occurs). However, to determine a similar heuristic between scenes would require too many assumptions to be considered a robust and fair evaluation configuration.

\begin{figure}[]
    \centering
    \includegraphics[width=0.5\linewidth]{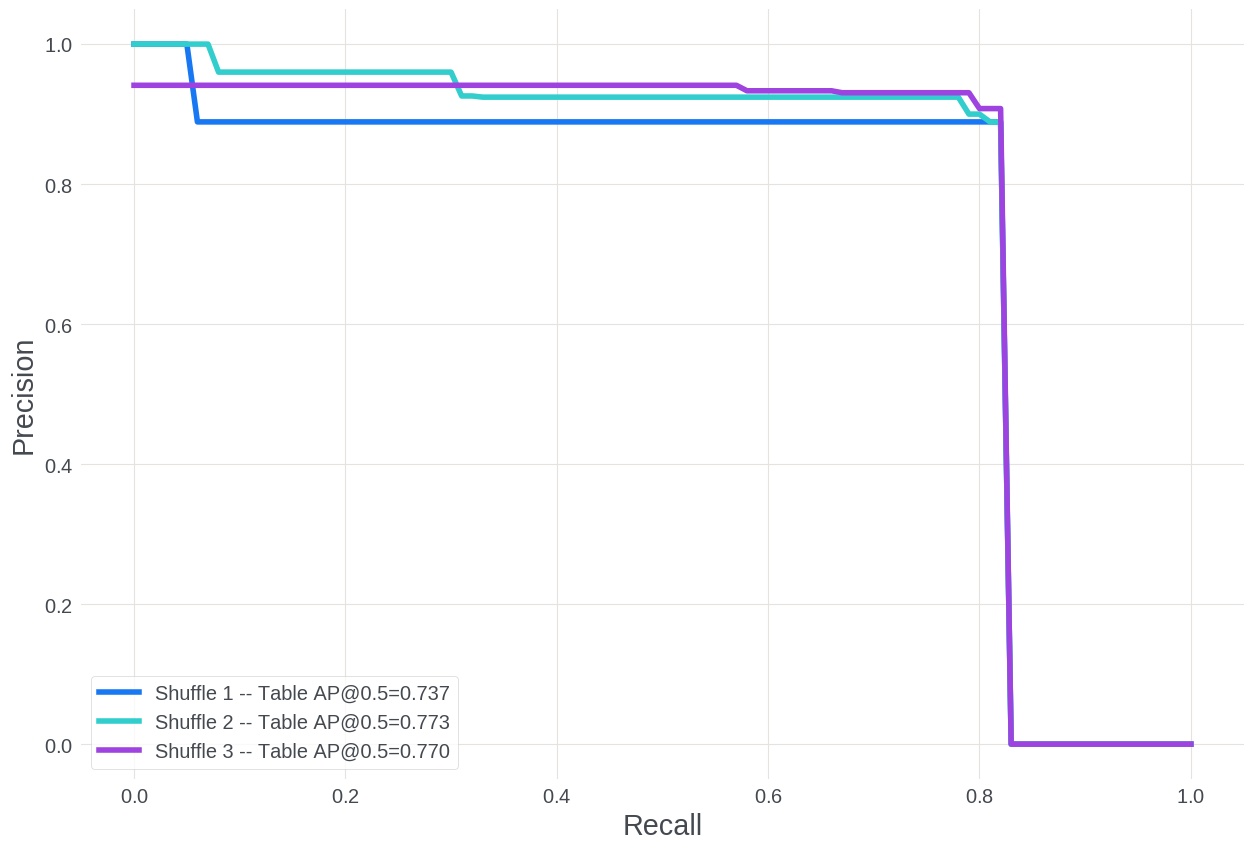}
    \caption{Precision-recall variance with scene order. The precision-recall curves are plotted for \METHOD{}'s predictions of the \textit{table} class on the same 10 scenes, however the order of those scenes is shuffled for each evaluation. The inability sort predictions across scenes leaves the AP@0.5IoU metric sensitive to the order that scenes are evaluated.}
    \label{fig:prec-rec}
\end{figure}

\begin{table}[]
    \centering
    \caption{Illustration of how average precision is negatively affected by the inability to sort across scenes. Two idential sets of detections are produced by detectors 1 and 2. Detector 1 outputs an absolute measure of confidence allowing for sorting across scenes. However, it is only possible to determine the relative confidence of predictions within a scene for detector 2. This results in a lower AP, as there is no opportunity to rank good predictions from scene B above bad predictions from scene A. We assume there are 3 GT entities in each scene for AP and F1 computation.}
    \begin{tabular}{c| cccccc | cccccc}
     & \multicolumn{6}{c|}{Detector 1} & \multicolumn{6}{c}{Detector 2} \\
     & \multicolumn{6}{c|}{w/ absolute conf.} & \multicolumn{6}{c}{relative conf. only} \\
     \hline \hline
     Scene    & A & B & B & A & B & A & A & A & A & B & B & B \\
     Certainty & high & high & high & med. & low & low     & - & - & - & - & - & - \\
     Success & 1 & 1 & 1 & 0 & 0 & 0    & 1 & 0 & 0 & 1 & 1 & 0 \\
     \hline
     AP & \multicolumn{6}{c|}{0.5} & \multicolumn{6}{c}{0.34} \\
     F1 & \multicolumn{6}{c|}{0.5} & \multicolumn{6}{c}{0.5}
    \end{tabular}
    \label{tab:pr-ranking}
\end{table}

To further illustrate this point, we consider an evaluation setup where we use prediction order within scenes as a proxy for relative certainty, without sorting across scenes. In Figure~\ref{fig:prec-rec} we show precision-recall curves computed over 10 scenes from \DatasetName{} validation set using the assumption. Importantly, each curve on this graph are the \textit{same detections on the same scenes}, but with the scenes simply evaluated in a new, random order each time. Not only is the resulting metric variant with the order of scenes, but low certainty predictions at the end of ascene's predictions may appear earlier in the ranked pool of detections than high certainty predictions from another scene. If these are incorrect, they will arteficially lower the precision achievable, and in turn lower the average precision for a method. A toy example of this is included in Table~\ref{tab:pr-ranking}.

For these reasons, in the main paper we choose to use a F1-score-based metrics to evaluate detection performance. These are not sensitive to ordering as also illustrated in Table~\ref{tab:pr-ranking}.

\begin{figure}[t]
    \centering
    \includegraphics[width=\columnwidth]{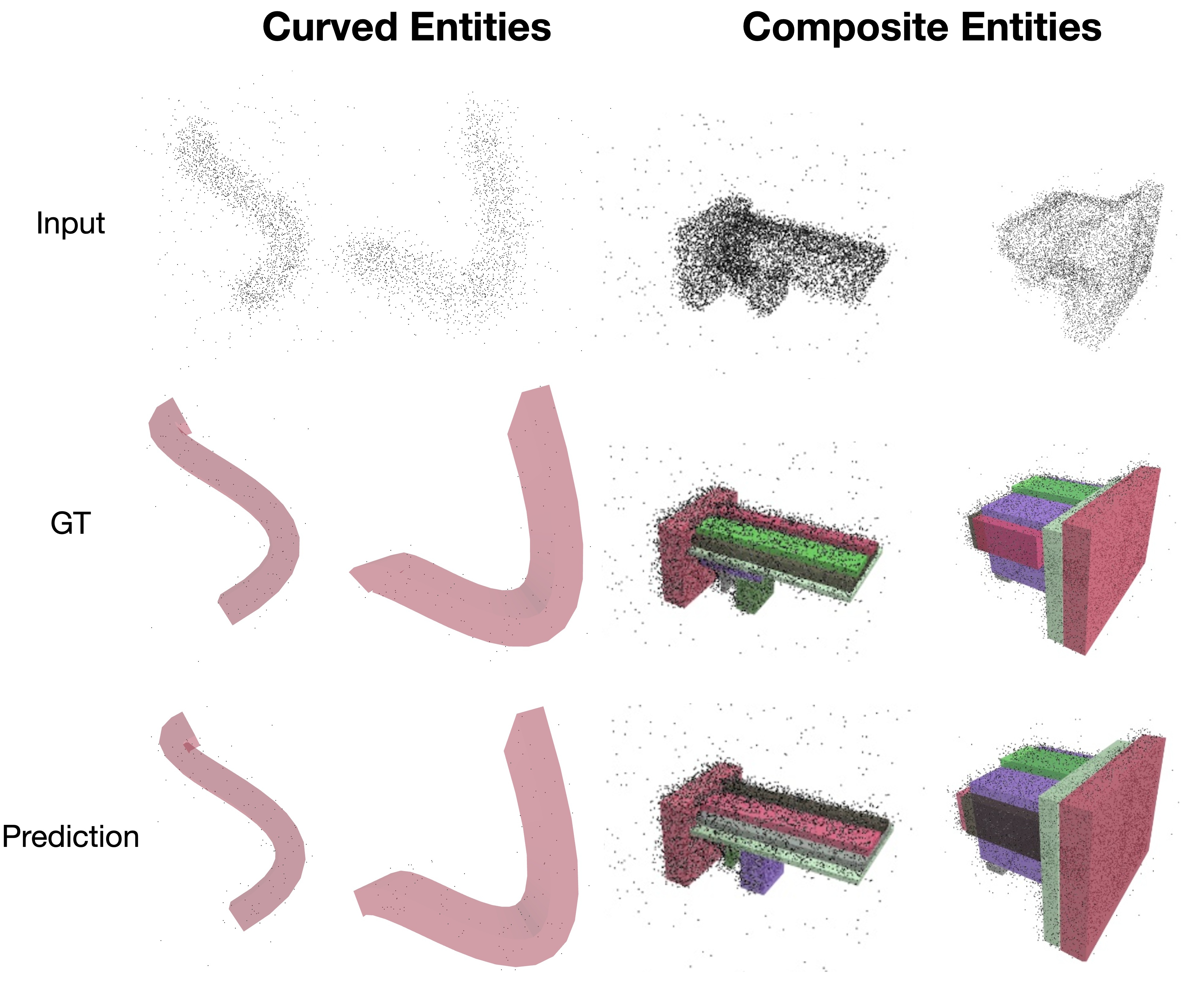}
    \caption{Non-planar wall geometry extensions to \METHOD. Examples of input point clouds (top row), the prediction 3D shape (middle row),
and ground truth wall shape (bottom row). (left) Examples of Bezier parameterisation for curved walls. (right) Results for wall primitive compositions. We observe that both simple extensions to the parameterisation of the walls can be accurately described and predicted by SceneScript.}
    \label{fig:ext_fuyang}
\end{figure}
\begin{figure}[ht]
    \centering
    \includegraphics[width=0.9\columnwidth]{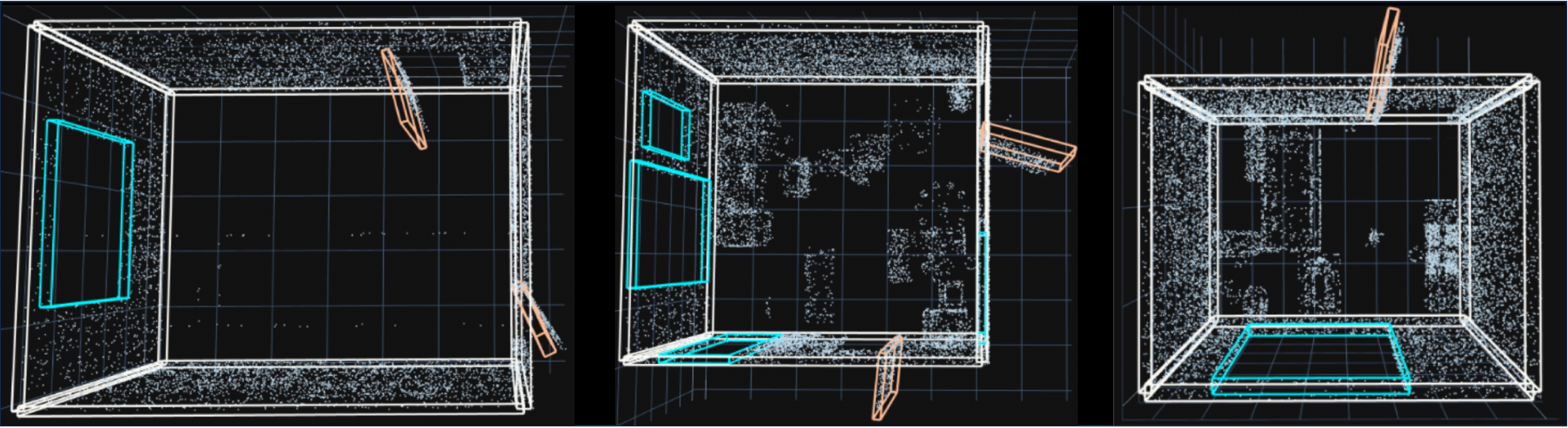}
    \caption{Results for detecting door state estimation. We visualize the predicted layout on top of the input point cloud.}
    \label{fig:ext_door_state}
\end{figure}

\section{Further Extensions of SceneScript}
\label{app:extensions}
\subsection{Extending \METHOD~with Curved Entities}

In previous layout estimation work, such as \cite{furukawa2009reconstructing,liu2019planercnn},
methods leverage a planar assumption
and use robust estimation procedures custom tailored to planar primitive fitting.
Extending such methods to more complex, non-planar entities is non-trivial and requires significant effort.
In contrast,
we show that our structured language-based approach makes it straightforward to extend to curved walls,
as for example extruded Bezier curves~\cite{beziercurves},
simply by defining a new \texttt{make\_curved\_wall} command. 

The curved wall command is a simple Bezier parametrisation
consisting of 4 additional parameters:
the $x,y$ values of the 2 control points
that define the wall curvature.
Explicitly, our planar wall command changes to: 
\begin{lstlisting}[language=StructuredLanguage]
make_curved_wall: a_x, a_y, a_z, b_x, b_y, b_z, c1_x, c2_y, c2_x, c2_y, height, thickness
\end{lstlisting}
where $c_{1_x}, c_{1_y}, c_{2_x}, c_{2_y}$ are the Bezier control points.

We generate a synthetic curved walls dataset to train \METHOD.
Example Bezier walls with a qualitative evaluation are in Figure~\ref{fig:ext_fuyang} (left).
The predictions are nearly indistinguishable compared to ground truth,
indicating that our method can learn to predict such complex primitives.

In a synthetic test bed, we evaluate the capability of our model to infer the control points of walls parameterized on extruded Bezier curves. Quantitative results are shown in Table~\ref{tab:quantitative_eval_bezier_cuves}.

\begin{table}[]
\centering
\caption{Quantitative assessment of the reconstruction of curved walls using extruded Bezier curves as parameters. Token accuracies gauge performance based on a tokenized 1D sequence of the structured language, allowing for a specified slack of +/- N tokens. The IOU is calculated by comparing the interpreted geometry with the GT geometry. We achieve virtually error-free results indicating efficient interplay between parameterization and modelling capability of our method.}
\begin{tabular}{c|c|c}
     Token Acc. Slack 1 & Token Acc. Slack 3 & IOU \\ 
    \hline
       0.993 & 1.0 & 0.990 \\
\end{tabular}
\label{tab:quantitative_eval_bezier_cuves}
\end{table}

\subsection{Extending \METHOD~to Compositions of Wall Primitives}

To demonstrate the extensibility of \METHOD's structured language,
and similarly to the reconstruction of object primitives explored in the main paper, we demonstrate representing complexly shaped walls
as compositions of cuboids.
We define a new parametrization for this class of walls as follows:
\begin{lstlisting}[language=StructuredLanguage]
make_wall: a_x, a_y, ...
make_wall_prim: pos_x, pos_y, pos_z, size_x, size_y, size_z
\end{lstlisting}
where the \texttt{make\_wall\_prim} command describes a cuboid
to be composed with its parent wall entity.
We added such cuboid compositions to a base wall
in Figure~\ref{fig:ext_fuyang} (right).
In this proof-of-concept,
the results of Table~\ref{tab:compositewalls} clearly demonstrate the ability of the network
to infer compositions of cuboids on base walls
only from a noisy surface point cloud.

\begin{table}[]
    \centering
    \caption{Correctly predicted parameters of composite walls as a percentage . Slack $n$ indicates estimation of composite wall parameters within bounds of $n*5cm$.}
    \begin{tabular}{@{}cc||*{3}{c|}}
  \multicolumn{1}{c}{}  &   &\multicolumn{3}{|c|}{Occlusion Levels}\\
    \multicolumn{1}{c}{} & & No    & Light    &High       \\ \hline \hline
    \multirow{3}*{\rotatebox{90}{Slack}}  
   & 1 & 99.6 & 95.9   & 92.6    \\ 
   & 3 & 99.9    &96.4 & 93.3     \\ 
   & 5 & 100    & 98.2    & 95.5   \\ 
   \end{tabular}
    \label{tab:compositewalls}
\end{table}
  
\subsection{Extending \METHOD~to Object States}

Yet another simple extension to our \METHOD~language allows us to represent door states, w.r.t their opening configuration. For this, we simply change the original door representation to include a list of parameters that define door state as follows:

\begin{figure}[t]
    \centering
    \includegraphics[width=\columnwidth]{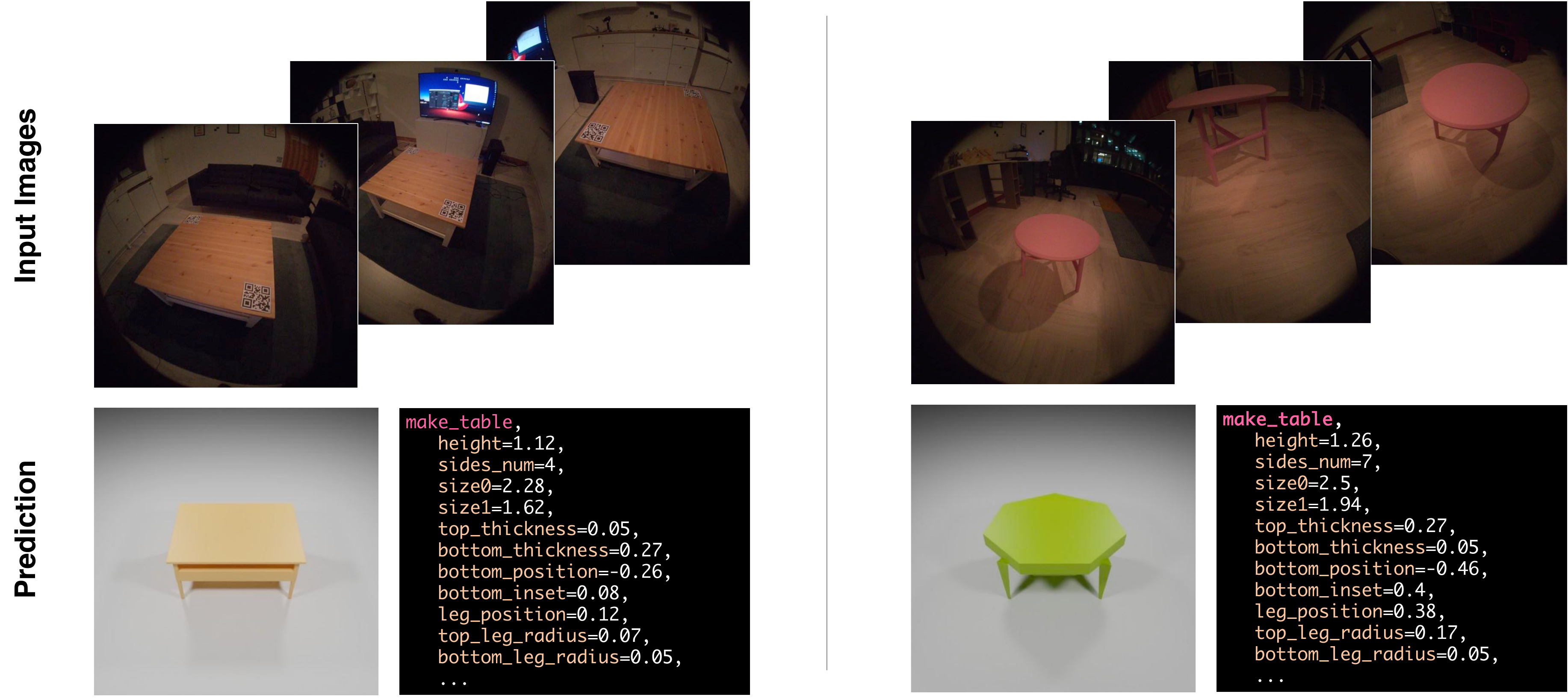}
    \caption{Two real-world inferences based on a Blender Geometry Node [3] obtained online. Input RGB images are recorded on an Aria device. We visualize a subset of the predicted language as well as the geometry obtained by inputting that prediction into the Geometry Node.}
    \label{fig:ext_blender}
\end{figure}

\begin{lstlisting}[language=StructuredLanguage]
make_door: id, wall_id, pos_x, pos_y, pos_z, width, height, open_degree, hinge_side, open_direction
\end{lstlisting}
\texttt{hinge\_side} represents which side of the door the hinge is on, \texttt{open\_direction} determines whether the door opens into the room or outside of the room, and \texttt{open\_degree} is the angle of the opening.
In Figure~\ref{fig:ext_door_state} (second),
we qualitatively demonstrate object state estimation.
We annotated our doors
with a new command parameterisation
extended by door hinge position, wall opening side and opening angle.
As with our other extensions,
our model is able to handle this situation without issue.
This small GT language extension demonstrates effective state estimation
while the input and network architecture remain unchanged.

\subsection{Extending \METHOD~to Blender Parametric Object Models}

Parametric modelling offers detailed high-quality geometry representations along with interpretability and editability by design~\cite{jones2020shapeassembly,jones2021shapemod,jones2022plad,pearl2022geocode}. 
The Blender community offers readily accessible Geometry Nodes of diverse object categories as a procedural language.
We investigate the use of a particular Geometry Node for tables~\cite{mrBash2023Tables}.
Not only can we directly incorporate this parametric model
into our \METHOD~language,
but we can also use it to generate data
by randomly sampling its parameters
similar to~\cite{pearl2022geocode}. 

We design a simple proof-of-concept experiment where we render synthetic RGB images of random tables, composite them on a random image background, and learn to predict the ground truth Blender procedural language.
In Figure~\ref{fig:ext_blender}, we demonstrate two real-world inferences of tables using this language, showing our method is capable of predicting reasonable parameters to reconstruct these tables. Interestingly, in the second example the model predicts a high \texttt{sides\_num} to approximate the circular tabletop, which was not on the training set.

\end{document}